\documentclass[journal]{IEEEtran}

\usepackage[colorlinks]{hyperref}
\usepackage{amsmath}
\DeclareMathOperator*{\argmax}{argmax}
\usepackage[numbers]{natbib}
\usepackage{algorithmic}
\usepackage{graphicx}
\usepackage{textcomp}
\usepackage{xcolor}
\usepackage{multirow}
\usepackage{subcaption}
\usepackage{colortbl}
\usepackage[table]{xcolor}
\usepackage{tikz}
\usepackage{amssymb}
\usepackage{amsmath}
\usepackage{makecell}

\definecolor{class2}{HTML}{1F77B4}   
\definecolor{class3}{HTML}{AEC7E8}   
\definecolor{class11}{HTML}{FF7F0E}  
\definecolor{class12}{HTML}{FFBB78}  
\definecolor{class14}{HTML}{2CA02C}  
\definecolor{class16}{HTML}{98DF8A}  
\definecolor{class17}{HTML}{D62728}  
\definecolor{class29}{HTML}{FF9896}  
\definecolor{class30}{HTML}{9467BD}  
\definecolor{class31}{HTML}{C5B0D5}  
\definecolor{class33}{HTML}{8C564B}  
\definecolor{class34}{HTML}{C49C94}  
\definecolor{class36}{HTML}{E377C2}  
\definecolor{class41}{HTML}{F7B6D2}  
\definecolor{class44}{HTML}{7F7F7F}  
\definecolor{class45}{HTML}{C7C7C7}  
\definecolor{class46}{HTML}{BCBD22}  

\ifCLASSINFOpdf
\else
\fi
\begin{document}
		%
		\title{DiMoP: Diffusion-Driven Motion Representation Learning With Frame-Level Pseudo-Classification for Skeleton-Based Action Recognition}
		%
		%
		%
		
		\author{Shanaka~Ramesh~Gunasekara,~\IEEEmembership{Student,~IEEE}, Wanqing Li$^{*}$,~\IEEEmembership{Senior Member,~IEEE}, Nikalal Kaldera,~\IEEEmembership{Student,~IEEE}, Philip Ogunbona,~\IEEEmembership{Senior Member,~IEEE},  Jack Yang,~\IEEEmembership{Senior Member,~IEEE.}
			\thanks{$^{*}$ corresponding author } 
			\thanks{Shanaka Ramesh Gunasekara, Wanqing Li, Philip Ogunbona and Jack Yang are with the Advanced Multimedia Research Lab, University of Wollongong, Australia.}}
		
		%
		%

	\markboth{IEEE Transactions on Biometrics, Behavior, and Identity Science}%
	{Shell \MakeLowercase{\textit{et al.}}: Bare Demo of IEEEtran.cls for IEEE Journals}
	%



	\maketitle
	
	\begin{abstract}
		Robust skeleton-based action recognition requires representations that capture a wide spectrum of motions, from subtle to moderate and strong ones. Existing methods often focus on strong motions. This paper introduces DiMoP, a masking- and diffusion-driven motion representation learning method with frame-level pseudo-classification to explicitly learn the distribution of joint motions rather than regressing deterministic coordinates, as existing methods often do. By diffusing masked joints with progressive noise and denoising them conditioned on visible joints, DiMoP learns through controllable noising and denoising processes, enabling uniform learning of weak, moderate, and strong dynamics. To enable the masking-based generative diffusion learning with a discriminative capability,  a pseudo-frame classifier is proposed that enforces the learning towards sequence-consistent and temporally coherent pseudo-labels without manual annotations. Together, these strategies provide a principled mechanism for joint generative and discriminative motion modeling. DiMoP achieves state-of-the-art performance across NTU RGB+D 60/120, and PKUMMD, including a 1.1 percentage point gain over prior works on NTU RGB+D 120 with the cross-subject protocol. The code is available at 
		\href{https://github.com/ShanakaRG/DiMoP-Diffusion-Driven-Motion-Representation-Learning-for-Skeleton-Based-Action-Recognition}{DiMoP.}
	\end{abstract}
	

	%
	\IEEEpeerreviewmaketitle

	\section{Introduction}
	\label{sec:introduction}
	
	Human action recognition is required in many diverse applications, including video surveillance, autonomous driving, robotics, healthcare monitoring, and industrial safety. It enables the detection of hazardous events, abnormal behaviors, and work-efficiency patterns. Compared with RGB-based methods~\cite{sltnNet,d3d}, skeleton data, obtainable from CCTV systems or commodity 3D cameras, has emerged as an efficient and robust alternative due to its lower computational cost and reduced sensitivity to environmental variations~\cite{wang01}. 
	
	There has been significant progress in supervised action recognition~\cite{blockgcn,shanaka01,shanaka02,Chen2021a,3mFormer,Bruno}. Despite this progress, performance remains highly dependent on large amounts of annotated training data. Collecting and annotating extensive datasets in the real-world is labor- and time-intensive. Consequently, unsupervised and self-supervised methods have emerged as promising alternatives~\cite{shanaka03,MacDiff,Wenhan01,actionlet,aimclr,Bulat}. 
	
	In self-supervised learning (SSL), learned representations are required to capture the intrinsic structure and distribution of the data without annotations to enhance downstream performance. To this end, various pre-training approaches are employed to learn effective representations. These approaches can be broadly categorized into two groups: contrastive methods~\cite{actionlet,aimclr,shanaka03,skelemixclr} and generative methods~\cite{Wenhan01,mao01,SkeletonMAE}.
	
	Contrastive approaches employ various transformations to approximate the data distribution in the real world. Typically, handcrafted transformations, such as random cropping, scaling, rotations, and temporal jittering, are used to generate multiple views of the same instance, thereby enhancing data diversity~\cite{aimclr,skelemixclr}. However, these operations provide only limited stochasticity and fail to simulate the true probabilistic distribution of motion dynamics, ultimately oversimplifying the complex and uncertain behaviors inherent in skeletal movements, particularly the subtle movements. 
	
	Recent generative methods typically adopt the Masked Autoencoder (MAE)~\cite{mae} framework for representation learning by reconstructing masked joints or body parts from unmasked or visible joints~\cite{Wenhan01,mao01,SkeletonMAE}. In these methods, random~\cite{MacDiff,Wenhan01} or sample-dependent masking~\cite{mao01} is often used. Although this design effectively models the relationship between unmasked and masked parts and captures their dynamics, \textit{the direct prediction of masked joints from unmasked joints often fails to learn subtle dynamics due to the simultaneous coexistence of weak and strong dynamics}, leading to ineffective modeling of subtle motion in the presence of highly complex and dynamic motion patterns in the real world and, hence, compromising the generalization of the learned representations.

	To improve the learning of subtle motion and the discriminative power of the learned representation, this paper proposes a framework for \textbf{Di}ffusion-Driven \textbf{Mo}tion Representation Learning with Frame-Level \textbf{P}seudo-classification, referred to as DiMoP. DiMoP is a joint diffusion-based generative and unsupervised discriminative learning model. First, DiMoP randomly selects joints and gradually diffuses them with controlled Gaussian noise, followed by denoising conditioned on the representation of visible joints. This progressive diffusion trajectory compels the model to learn subtle motion variations in the early steps and high-energy dynamics in the later steps, providing a principled mechanism that treats all motion scales equally: subtle, moderate, and strong. Second, a frame-based pseudo-classification module is introduced to enhance the discriminativeness of the learned motion representation. The pseudo-classification module transforms a sequence of uniformly distributed random numbers of the same length as the action sequence into a frame-based pseudo-action label, conditioned upon the learned motion representation of the unmasked joints or parts. Note that we refer to the module as \textit{a pseudo-frame-based classifier or a pseudo-classifier} because the actual action label is unknown, and the classifier aims to enforce the learned representation to achieve consistent and temporally coherent frame-based labels for the entire sequence. 
	
	
	The key contributions of this work are summarized as follows:
	\begin{enumerate}
		\item A generative-discriminative framework, DiMoP, is proposed. DiMoP replaces deterministic masking with a progressive noise–denoise process, enabling the model to fully capture the joint motion distribution and effectively learn the subtle, moderate, and strong dynamics. These capabilities are often lacking in existing masking-based methods. 
		\item The introduction and integration of a pseudo-frame-based classifier module into DiMoP enhance the discriminative power of the learned motion representation. This mechanism effectively enforces explicitly discriminative learning in a diffusion-based generative model. 
		\item Extensive experiments were conducted on four benchmark datasets to demonstrate that the proposed generative–discriminative design consistently yields state-of-the-art or competitive performance, thereby validating its effectiveness for skeleton representation learning.
	\end{enumerate}

	
	
	
	
	\section{Related Work}
	
	\label{sec:related}
	
	This section briefly reviews recent advances in supervised and self-supervised skeleton-based action recognition.
	
	
	\subsection{Supervised Methods}
	
	In early-stage works, CNN~\cite{Wang05,wang03,CNN1}  and RNN~\cite{Yong0123,LSTM01} architectures were often employed for action recognition, while more recent studies have been dominated by GCN~\cite{blockgcn,degcn,Chen2021a,shanaka02} and transformer-based architectures~\cite{skateformer,3mFormer} due to their promising performance.
	
	Graph convolutional networks (GCNs) for skeleton-based action recognition were introduced by \textit{Yan et al.} in the ST-GCN framework~\cite{Yan01}. Subsequently, various techniques have been explored to improve model accuracy and robustness, including the use of adaptive data-driven topologies~\cite{blockgcn,LiKBS02}, the expansion of spatial and temporal receptive fields~\cite{Cheng2017, LiKBS01}, adaptive feature aggregation~\cite{shanaka01,shanaka02,KilicKBS01}, and prompt-based learning strategies~\cite{gap}.
	
	In ST-TR~\cite{Chiara01}, vision transformers are employed for skeleton action recognition through the introduction of distinct spatial and temporal attention modules. Subsequent works have explored hybrid transformers by integrating transformers with GCNs~\cite{jtg,hou01}  or temporal convolutions~\cite{Qiu01}, along with attention modules and developing parameter-efficient models~\cite{skateformer}. Recent studies have demonstrated that mask modeling is an effective approach for training vanilla transformer networks. The proposed DiMoP adapts a vanilla ViT~\cite{Vit} as the encoder and extends transformers to function as both the denoising decoder and the pseudo-frame-based classifier.

	\subsection{Unsupervised Methods}
	Recent self-supervised approaches have employed contrastive~\cite{actionlet,aimclr,crossclr,shanaka03} and generative methods~\cite{MacDiff,Wenhan01,SkeletonMAE,motionbert} to learn meaningful representations from unlabeled data, thereby improving performance in multiple downstream tasks.
	
	In contrastive learning, representations are learned by distinguishing between positive and negative pairs, which are typically generated using transformation functions that enhance the diversity of training samples. For instance, SkeletonCLR~\cite{crossclr} employs shear and crop transformations. Recent studies demonstrate that stronger transformations, those that incorporate rich semantic information, can significantly improve generalizability and reduce the gap with fully supervised methods. Building on this idea, AimCLR~\cite{aimclr} introduces extreme transformations to generate novel movement patterns, employing four spatial transformations (shear, spatial flip, rotation, axis mask), two temporal transformations (crop, temporal flip), and two spatio-temporal transformations (Gaussian noise, Gaussian blur). In contrast, ActCLR~\cite{actionlet} applies extreme transformations only to static body parts to preserve critical motion-related information. Despite the introduction of various data transformations, many fail to effectively simulate the complex motion dynamics, limiting their ability to capture the full range of movement variations.
	
	
	Masked Autoencoders (MAEs)~\cite{mae} have recently been actively adopted in 3D action representation learning. Transformer-based MAEs were reported in SkeletonMAE~\cite{Wenhan01} to reconstruct randomly masked joints. Specifically, visible joints are mapped to latent representations by the encoder, after which masked joints are replaced with fixed embeddings referred to as learnable tokens. These tokens, along with the latent representations of visible joints, are then fed into the reconstruction decoder to recover the masked joint coordinates. Subsequent works explored different masking operations or reconstruction targets to improve representation learning. For instance, in MAMP~\cite{mao01}, highly mobile joints were masked with a higher probability, and the original motion was reconstructed starting from learnable tokens. However, because every masked joint is represented by the same learnable token, the model’s ability to capture the variability inherent in human motion is limited.  In contrast, in DiMoP, diffused joints are used to replace the conventional masked patches, thereby encouraging the network to learn a probabilistic mapping from noisy to clean joint configurations. 
	
	Recently proposed MacDiff~\cite{MacDiff} employs a masked conditional diffusion framework where the denoiser is conditioned via global Adaptive Layer Normalization (AdaLN). The global representation, obtained by pooling the local features of visible joints from the semantic encoder, is used to uniformly scale and shift all feature channels in each denoiser layer. This global conditioning drives the denoising of the entire diffused skeleton, updating both visible and masked joints under the same modulation. Such global feature modulation provides coarse, sequence-level conditioning but lacks joint- or spatial-specific distinction. In contrast, DiMoP introduces a structured partial-state diffusion that diffuses only masked joints while keeping visible joints as clean anchors throughout the denoising. Instead of global AdaLN modulation, DiMoP performs token-level cross-conditioning via cross-attention; queries from noisy (masked) joints attend to keys/values from visible-joint embeddings, enabling localized, motion-aware guidance that MacDiff’s global feature modulation cannot achieve.


	
	\section{Proposed Method}
	\label{Ch6_proposed_method}
	
	
	\subsection{Overview}
	
\begin{figure*}
		\centering
		\includegraphics[width=0.8\linewidth]{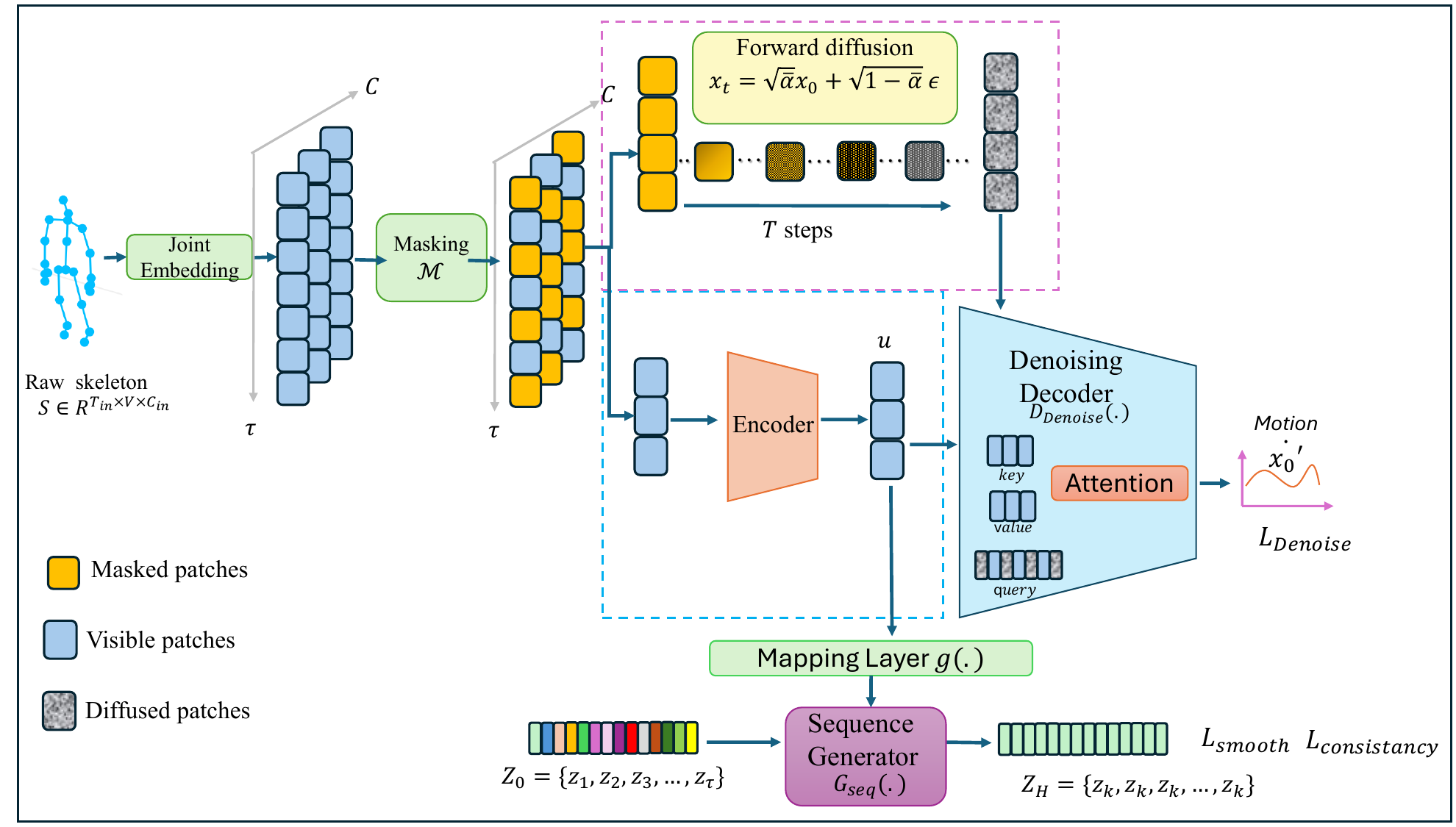}
		\caption [Overview of DiMoP framework]{Overview of the proposed DiMoP framework. A given skeleton sequence is embedded into tokens, and random masking is applied. The encoder $E(.)$ maps visible joints to the latent space. The decoder $D_{Denoise}$ uses the latent representation of the visible joints as a condition to denoise the diffused masked joints and recover the original motion. The $G_{psu}$ transforms a random sequence of distinct numbers into a homogeneous sequence to assign frame-wise pseudo-labels to enforce the discriminative power of the encoder.}
		\label{fig:diff_mask}
	\end{figure*}

	Figure~\ref{fig:diff_mask} depicts the network architecture of the proposed method, DiMoP, which takes a raw skeleton sequence $S$ as input, where $S \in \mathbb{R}^{T_{in} \times V \times C_{in}}$, $T_{in}$, $V$ and $C_{in}$ indicate the number of input frames, the number of joints, and the number of input channels, respectively. Following common practice in vision transformers~\cite{Vit}, a linear embedding layer is utilized to map the input joint coordinates into joint embeddings.
	
	
	The DiMoP network consists of four modules: an encoder $E(\cdot)$, a denoising decoder $D_{\text{Denoise}}(\cdot)$, a mapping layer $g(\cdot)$, and a pseudo-classifier $G_{\text{psu}}(\cdot)$, where all except the mapping layer follow the standard transformer architecture~\cite{attention}.

	Given a skeleton sample in the embedding space $x_0 \sim q(x_0)$ (where the subscript 0 denotes the original sample), a masking operation $\mathcal{M}$ first decomposes $x_0$ into two parts:  the masked joints $x_0^m$ and the visible joints $x_0^{v}$. In DiMoP, the term masking refers to an indexing operation that separates the input joints into visible and masked subsets rather than replacing the masked joints with zero or a fixed value. The visible joints $x_0^{v}$ are passed through the encoder $E(\cdot)$ to produce latent representations, while the masked joints $x_0^m$ retain their original values and go through the forward diffusion process. Specifically, the masked subset is progressively noised according to Equation~\ref{eq6_1}, generating $x_t^m$   at diffusion step $t$. The denoising decoder $D_{\text{Denoise}}(\cdot)$ then takes the diffused masked joints and the latent representation of the visible joints as inputs, ultimately recovering the motion of the original sample $x_0$.
	
	
	\begin{align}
		x_t = \sqrt{\Bar{\alpha}} x_0 +  \sqrt{(1-\Bar{\alpha}_t)} \epsilon, 
		\label{eq6_1}
	\end{align}
	where, $x_t$ denotes the diffused sample at $t$, $\alpha_t = 1- \beta_t$, and $\Bar{\alpha}_t = \prod_{i=1}^t \alpha_i$ ($\beta_1,...,\beta_T$ are predefined variance schedules for $T$ steps),  $\epsilon \sim \mathcal{N}(0,\textbf{I})$ and $\textbf{I}$ is an identity matrix. 
	
	Simultaneously, the encoder output $u$ is converted by a mapping layer into a conditioning signal for $G_{psu}(\cdot)$, which transforms a randomly generated sequence into a homogeneous sequence via the pseudo-classifier. The denoising of masked joints and the pseudo-label generated by the classifier are used jointly to guide the encoder in enhancing its representational and discriminative capabilities. 
	
	After pre-training, only the joint embedding layer, encoder, and mapping layer are retained, with encoder and mapping outputs concatenated along the channel dimension for downstream inference.
	
	\subsection{DiMoP}

	\subsubsection{Encoder}
	An input sequence \(S \in \mathbb{R}^{T_{in} \times V \times C_{in}}\) is divided into 
	\(\tau = T_{in}/l\) non-overlapping temporal segments and embedded by \(g_1(\cdot)\), resulting in \(x \in \mathbb{R}^{\tau \times V \times C}\). Spatial and temporal positional embeddings are then added following the transformer design~\cite{attention}: \(x_0 = x + pe^s + pe^t\), where \(pe^s \in \mathbb{R}^{1 \times V \times C}\) and \(pe^t \in \mathbb{R}^{\tau \times 1 \times C}\). The resulting \(\tau \times V\) tokens are treated as the clean sample \(x_0\).
	
	A random mask with a ratio \(r\) separates \(x_0\) into visible joints \(x_0^v\) and masked joints \(x_0^m\), where \(\Gamma=(1-r)\tau V\) joints remain visible. A high masking ratio \(r=0.9\) was set to form a strong representation for learning and to reduce encoding time. The random mask is used as an unbiased denoising learning strategy rather than as a sensor-occlusion simulation. Although real skeleton noise can be joint-dependent, an explicit occlusion prior may over-emphasize frequently corrupted or highly mobile joints. Uniform masking instead enables all joints, including low-motion but discriminative ones, to contribute to conditional denoising, as supported by the masking ablation in Section~IV-D1.

	
	The visible joints map to the latent space via the encoder.
	\begin{equation}
		u = E\bigl(x_0^v\bigr) \in \mathbb{R}^{\Gamma \times C},
	\end{equation}
	where $u$ is the latent representation of $x_0^v$.
	This representation \(u\) is employed as the conditioning signal for both the denoising decoder and the pseudo-classifier.
	
	After pre-training, only the encoder \(E(\cdot)\) and the mapping layer $g(\cdot)$ are fine-tuned for downstream tasks.

	\subsubsection{Forward diffusion}
	
	In the forward diffusion process, only the masked joints are progressively diffused by repeatedly adding small amounts of Gaussian noise over \(T\) steps, as specified by Equation~\ref{eq6_1}. 
	
	The objective is to generate the masked joints through sampling, which is approximated by recursively drawing samples from  \(q(x_0^m \mid x_0^v)\), beginning with 
	\(x_T^m \sim \mathcal{N}(0, I)\). When the variance of the noise \(\beta_t\) at each step \(t\) is sufficiently small, the distribution  \(q\bigl(x_{t-1}^m \mid x_t^m, x_0^v\bigr)\) can be treated as Gaussian \cite{Jascha01}. Therefore, it can be approximated by a deep network.

	\subsubsection{Denoising Decoder}
	\label{6_decoder}
	
	The denoising decoder takes noisy masked joints $x_t^m$ as input and utilizes the latent representations of the visible joints $u$ as conditioning information. The noise level of each diffused patch is indicated by the timestep $t$, which is uniformly sampled from $[1,T]$ during training.
	
	Three decoder configurations are examined, differing in how attention is applied between visible latents and noisy tokens.
	
	\textit{Joint Decoder:} This configuration takes the full set of tokens (joints), $\tau\times V$, as input, which consists of (i) encoded visible joints and (ii) diffused mask joints placed at their corresponding indices (visible indices $idx_{visible}$ and mask indices 
	$idx_{mask}$ ). Queries, keys, and values are obtained through linear projections of this combined token set. However, by mixing visible and masked joints without explicit separation, there is a risk of interference between the clean conditional data and the noisy inputs.
	
	\textit{Cross Decoder V1:} In this variant, each decoder block employs a cross-attention layer where noisy joints first attend to the latent representation of visible joints. Specifically, queries are computed through a linear projection of the noisy joints, while keys and values are derived solely from the linear projections of the latent visible joints. This iterative process denoises each noise patch conditioned on the constant visible information. One drawback is that each noisy joint attends independently to the visible joints, without considering the interactions among other noisy joints.
	
	\textit{Cross Decoder V2:} This configuration takes the complete joint set as input but treats visible joints as a separate conditioning branch. Queries are computed through a linear projection of the entire joint set, while keys and values are derived solely from the linear projections of the latent visible joints. This design preserves the global context in the queries by allowing each joint to benefit from interactions with the complete set, while the exclusive conditioning on visible joints ensures a stable and reliable guidance signal during denoising.
	
	Combining $u$ and $x_t^m$ allows the model to learn the conditional distribution 
	$p(x_0^m \,|\, x_t^m, x_0^v)$, where $u$ acts as a learned prior guiding 
	structure-aware denoising. This formulation aligns noisy joints with plausible 
	motion manifolds, producing stable and discriminative representations. 
	
	Three decoder architectures are ablated in Section~\ref{ablation6}. In all scenarios, positional embeddings are added to the tokens (joints) before they are fed into the decoder. The model linearly projects the decoder output to generate the final prediction, which has the same shape as \( x_0 \).  The  $D_{Denoise}$ is only used to denoise the diffused masked joints during training and will not be used for downstream tasks. 
	
	\subsubsection{Pseudo-classifier}
	
	The pseudo-frame classifier serves as an auxiliary head that imposes a frame-level discriminative constraint. In particular, the pseudo-classifier enforces temporal coherence by mapping all frames of a sequence toward a consistent label space, thereby suppressing frame-level noise and compacting sequence-level representations. Although not tied to true action labels, these homogeneous pseudo-labels act as latent surrogates that enhance feature separability and strengthen discriminative power for downstream recognition.
	
	While diffusion learning primarily reconstructs motion dynamics, it may yield overly smooth representations; the pseudo-classifier mitigates this by enforcing temporal consistency within actions and separation across distinct motion phases, effectively introducing an implicit temporal clustering signal that enhances inter-class separability and intra-action coherence without real labels. Specifically, the pseudo-frame-based classifier \(G_{psu}(\cdot)\) is driven by a random sequence and generates a homogeneous sequence conditioned on the encoder output \(u\). 
	
	The random sequence consists of distinct random integers $Z_0 = \{z_1,z_2,z_3,...,z_i,...,z_\tau\} \in \mathcal{R}^{\tau\times 1}$, where each element \( z_i \in [1, \tau]\). Based on the encoder output \( u \), the classifier produces a homogeneous sequence $Z_H = \{z_k, z_k, z_k, \dots, z_k\} \in \mathbb{R}^{\tau \times 1},$ where  $z_k \in \{z_1,z_2,z_3,...,z_\tau\}$. The classifier is formally expressed as: $Z_H = G_{psu}(Z_0, u)$.
	
	\subsubsection{Training Objective}
	
	The model is pre-trained using a combination of denoising loss and the homogeneity of the label sequences produced by the pseudo-classifier. The homogeneity loss is assessed by measuring the temporal smoothness and consistency of the pseudo-labels across the frames.
	
	
	\textit{\textbf{Denoising loss:}} Unlike the standard denoising Diffusion Probabilistic
	Models (DDPM) approach~\cite{Jascha01}, which often employs $\epsilon$-prediction for noise estimation, the proposed DiMoP model directly reconstructs the clean sample through a denoising process. Both formulations are widely adopted in diffusion models~\cite{Jascha01}. In particular, DiMoP is trained to predict the motions of masked joints by denoising the noisy input, thus recovering the underlying motion distributions.
	
	The denoising objective is defined as:
	\begin{equation}
		\mathcal{L}_{Denoise} = \mathbb{E}_{t, x_0, \epsilon} \Bigl\| \dot{x}_0^m - D_{\text{Denoise}}\bigl(x_t^m, t, E(x_0^v)\bigr) \Bigr\|^2,
	\end{equation}
	where $\dot{x}_0^m$ is the motion of the original sequence $x_0$.

	\textit{\textbf{Temporal smoothness:}} 
	The temporal smoothness loss regularizes local frame-to-frame variations in the pseudo-classifier output. The output of $G_{psu}(\cdot)$ is first passed through a \textit{softmax} layer, and the frame-level pseudo-label sequence is obtained as
	$Z_H=\argmax(\text{softmax}(G_{psu}(\cdot)))$. Since $Z_H$ represents frame-wise pseudo-class assignments within the same action instance, abrupt transitions between adjacent frames are discouraged. To this end, a binary sequence $Z_B \in \{0,1\}^{\tau-1}$ is generated by comparing consecutive predictions in $Z_H$, where $Z_B[i]=1$ if the predicted value changes between frames $i$ and $i+1$, and $Z_B[i]=0$ otherwise. The smoothness loss is then defined as, 
	\begin{align}
		\mathcal{L}_{smooth} = \dfrac{1}{\tau-1} \sum_{i=1}^{\tau-1} Z_B[i].
		\label{Lsmmoth}
	\end{align}
	Therefore, $\mathcal{L}_{smooth}$ penalizes local temporal discontinuities and encourages smooth pseudo-label evolution. However, since it only measures adjacent transitions, it does not explicitly enforce sequence-level agreement among all frames.
	
	\textit{\textbf{Value consistency loss:}}
	To impose a global sequence-level constraint, a value consistency loss was introduced that encourages all frames to support the dominant pseudo-label estimated over the entire sequence. Unlike $\mathcal{L}_{smooth}$, which only penalizes local transitions, this loss directly promotes global agreement across all frame-level predictions.
	Specifically, for each candidate value $z_k$, its sequence-level vote is computed as
	\begin{equation}
		vote_{k} = \sum_{r=1}^{\tau} p_{r,k},
	\end{equation}
	where $p_{r,k}$ denotes the probability of assigning frame $r$ to value $z_k$. The dominant pseudo-label is then selected as
	\begin{equation}
		z^{maj} = \argmax_{k}(vote_{k}).
	\end{equation}
	Finally, each frame is encouraged to agree with this dominant value:
	\begin{equation}
		\mathcal{L}_{consistency} = 1 - \dfrac{1}{\tau} \sum_{r=1}^{\tau} p_{r}^{maj},
		\label{lconsist}
	\end{equation}
	where $p_r^{maj}$ is the probability assigned to the majority value at frame $r$.
	Thus, $\mathcal{L}_{smooth}$ and $\mathcal{L}_{consistency}$ are complementary: the former regularizes local temporal continuity, while the latter enforces global sequence-level consensus.

	The total loss function for the proposed network is expressed as a combination of \(\mathcal{L}_{Denoise}\), the temporal smoothness loss \(\mathcal{L}_{smooth}\), and the consistency loss \(\mathcal{L}_{consistency}\),
	

	\begin{equation}
		\begin{aligned}
			\mathcal{L}
			&=
			\alpha\mathcal{L}_{Denoise}
			+
			\gamma\left(
			\mathcal{L}_{consistency}
			+
			\frac{\lambda}{\gamma}\mathcal{L}_{smooth}
			\right),
		\end{aligned}
		\label{6lfinal}
	\end{equation}
	where $\alpha$ and $\gamma$ balance the generative denoising and
	discriminative pseudo-classification objectives, respectively, while
	$\lambda/\gamma$ controls the contribution of local temporal smoothness
	relative to global sequence consistency. 
	
	
	\section{Experiments}
	
	\subsection{Datasets}
	
	The proposed DiMoP is evaluated on three commonly used large-scale datasets.
	
	\subsubsection{NTU RGB+D 60~\cite{shahroudy01}} NTU RGB+D 60 is a large-scale dataset for human action recognition that contains 60 action categories performed by 40 subjects, totaling 56,880 skeleton sequences. The dataset is evaluated using two standard protocols: cross-subject (X-Sub) and cross-view (X-View). In the X-Sub protocol, action sequences from 20 subjects are used for training, while sequences from the remaining subjects are reserved for testing. In the X-View protocol, training samples are captured from cameras 2 and 3, while testing is conducted on samples from camera 1.
	
	\subsubsection{NTU RGB+D 120~\cite{Liu1905}} NTU RGB+D 120 extends NTU RGB+D 60 by doubling the action categories from 60 to 120. The dataset comprises 114,480 skeleton sequences and is evaluated using cross-subject (X-Sub) and cross-setup (X-Set) protocols. In X-Set, sequences are divided into 32 setups based on camera distance and background, with half allocated for training. The unique camera viewpoints ensure diversity in positions, angles, and backgrounds. The dataset covers a broad range of actions, including in-office activities, object interactions, fine-grained single-person actions (which are subject to occlusions and noise), and mutual interactions.
	
	\subsubsection{PKUMMD~\cite{Liu1703}} PKUMMD is a large-scale dataset for 3D human action recognition, divided into two parts. Part 1 contains over 1,000 action instances across 51 classes performed by 66 subjects, while Part 2 includes approximately 20,000 instances under more challenging conditions, characterized by increased variability, occlusion, and overlap. 
	
	
	
	These datasets provide complementary evaluation settings with distinct degradation factors and purposes. NTU RGB+D 60 serves as a standard benchmark for comparison with existing self-supervised skeleton representation learning methods, covering viewpoint changes, inter-subject variation, intra-class diversity, and skeleton tracking noise. NTU RGB+D 120 evaluates scalability and robustness in a larger and more diverse setting, with 120 action classes, 106 subjects from 15 countries, wide age and height ranges, multiple camera setups, fine-grained actions, object interactions, and diverse skeleton variations. PKUMMD assesses generalization under more realistic sequence-level challenges, including long continuous actions, action transitions, temporal ambiguity, subject overlap, and occlusion. Collectively, these datasets evaluate DiMoP across standard benchmark comparison, large-scale recognition, and challenging settings, while covering action classes with subtle, strong, and mixed subtle–strong motion patterns.
	
	\subsection{Implementation}
	
	The encoder was adapted from MAMP~\cite{mao01}. The model takes a 300-frame-long skeleton sequence as input, which is then cropped and interpolated to 60 frames. The patch size $l$ is set to 2 to generate non-overlapping temporal segments, which facilitates $\tau=30$.
	
	The embedding dimension of 256 is assigned to all three networks, and each multi-head self-attention (MSHA) block in \(E(\cdot)\) and \(D_{\text{Denoise}}(\cdot)\) is configured with 8 heads. The encoder consists of eight layers, while the denoising decoder comprises five layers. The pseudo-classifier consists of cross-attention blocks. Before the first layer of each module, separate learnable spatial and
	temporal positional embeddings are added to the input embeddings. A masking ratio of 0.9 is applied.  For the forward diffusion process, a total of \(T = 300\) sampling steps are used, and a linear noise scheduling scheme~\cite{ho01} is adopted.
	
	A pre-training phase consisting of 400 epochs is conducted with a batch size of 32 on nine NVIDIA P100 GPUs.  The first 20 epochs serve as a warm-up stage to ensure stable training. During this phase, the learning rate is elevated from 0 to \(1 \times 10^{-3}\), and then gradually lowered to \(5 \times 10^{-4}\) using a cosine decay schedule.  The Adam optimizer is employed to update the model parameters. 
	
	\subsection{Comparison with State-of-the-art Methods}
	
	\subsubsection{Linear evaluation}
	
	During the linear evaluation, a linear classifier $\phi(.)$ is placed on top of the pre-trained encoder $E(.)$ and the mapping layer $g(.)$. Both $E(.)$ and $g(.)$ are kept frozen, while only $\phi(.)$ is optimized for action classification. The model was trained for 150 epochs with a batch size of 128 and an initial learning rate of 0.1. Table~\ref{tab6_1} compares the performance of DiMoP with state-of-the-art (SOTA) approaches. DiMoP outperformed the baseline MAMP~\cite{mao01} by 2.8 and 1.9 percentage points on the NTU RGB+D 60 dataset using X-view and X-sub evaluations, respectively. Additionally, it outperformed MacDiff~\cite{MacDiff} by 1.6 percentage points on the NTU RGB+D 120 X-set evaluation.

	\begin{table*}[t]
		\begin{center}
			\caption[Linear evaluation of DiMoP ]{Linear evaluation performance of DiMoP versus state-of-the-art methods on the NTU RGB+D 60, NTU RGB+D 120, and PKUMMD datasets. The term ``3s-'' denotes the ensemble results from the joint (J), bone (B), and motion (M) streams. \textbf{Bold} and \underline{underlined} entries indicate the best and second-best performances, respectively.} 
			\label{tab6_1}
			\begin{tabular}{l | l | c|  l |c |c |c |c }
				\cline{1-8} 
				Models & Venue &Stream &\multicolumn{2}{c|}{NTU 60}  & \multicolumn{2}{c|}{ NTU 120} &\multicolumn{1}{c}{ PKU MMD} \\
				\cline{1-8} 
				&     &   &  X-sub  &  X-view  &  X-sub &  X-set  & Part I \\
				
				\cline{1-8} 
				{\textit{Contrastive Methods} }   & & & & & & &   \\
				ISC~\cite{thoker01}       & MM'21 &  J   & 76.3 & 78.6 & 67.9  & 67.1 & 80.9 \\
				GL-Transformer~\cite{glformer}  & ECCV'22&J  & 76.3 & 83.8 & 66.0 & 68.7& -  \\
				PSTL~\cite{pstl}        & AAAI'23 &  J  & 77.3 & 81.8  &  69.2 & 67.7 & 88.4 \\
				CPM~\cite{cpm}      & ECCV'22     &   J & 78.7 & 84.9 & 68.7 & 69.6 & - \\
				USDRL~\cite{USDRL}          & AAAI'25& J   & 85.2  & \underline{91.7}    & 76.6  & 78.1  \\
				U-FEFP~\cite{chuankun02}    & TCSVT'25& J & 86.7& 91.2& 78.23& 79.6 & \underline{92.9}\\
				3s-SkeletonCLR ~\cite{crossclr}  & CVPR'21 &J+B+M  & 75.0 & 79.8     &  60.7      &  62.6  & - \\ 
				3s-CrosSCLR ~\cite{crossclr}    & CVPR'21 &J+B+M   & 77.8 & 83.4     & 67.9   & 66.7 & 84.9 \\
				3s-AimCLR ~\cite{aimclr}        & AAAI'22 &J+B+M   & 78.9 & 83.8     & 68.2   & 68.8& 87.8 \\
				3s-PSTL~\cite{pstl}             & AAAI'23 &J+B+M   & 79.1 & 83.8  & 62.2 &70.3 & 89.2  \\ 
				3s-CPM~\cite{cpm}     & ECCV'22               &J+B+M   & 83.2 & 87.0  & 73.0 & 74.0 & - \\
				3s-ActCLR ~\cite{actionlet}     & CVPR'23 &J+B+M    & 84.3  & 88.8     & 74.3   & 75.7 & - \\
				3s-STJD-CL~\cite{shanaka03} &TBIOM'25 & J+B+M & 85.9 & 90.0 & 77.1 & 79.3 & - \\
				
				\cline{1-8} 
				{\textit{Generative Methods:}}& & & & & & &   \\
				SkeletonMAE ~\cite{Wenhan01} &ICMEW'23 & J & 74.8 & 77.7 & 72.5 & 73.5 & 82.8  \\
				MAMP ~\cite{mao01} & ICCV'23 & J & 84.9 & 89.1 & 78.6 & 79.1& 92.2  \\
				S-JEPA & ECCV'24 & J &85.3 &89.8 &\underline{79.6} &79.9 &92.2 \\
				MacDiff~\cite{MacDiff} & ECCV'24 & J  & \underline{86.4} & 91.0 & 79.4 & 80.2 & 92.8 \\
				STJD-MP~\cite{shanaka03} & TBIOM'25 & J & 85.4 & 90.2& 79.1& \underline{80.4} & - \\
				\cline{1-8} 
				DiMoP(Ours)  & & J & \textbf{86.8}& \textbf{91.9}& \textbf{80.7}& \textbf{81.8}& \textbf{93.2}\\
				\cline{1-8} 
				
			\end{tabular}
		\end{center}
	\end{table*}

	The statistical reliability of DiMoP was further assessed using bootstrap confidence estimation on NTU RGB+D 60 X-sub and X-view. Test samples were resampled with replacement for 50 iterations, and 95\% confidence intervals were computed from the resulting accuracy distributions. On X-sub, 86.80\% accuracy was achieved by DiMoP with a 95\% CI of [86.53, 87.01], exceeding MAMP [84.27, 85.13] and MacDiff [85.93, 86.56]. On X-view, 91.90\% accuracy was obtained with a 95\% CI of [91.62, 92.13], also exceeding MAMP [88.79, 89.42] and MacDiff [90.61, 91.18]. These results indicate that the observed improvements are statistically significant.

	\subsubsection{Supervised fine-tune evaluation}
	\label{supervse}
	
	A classification head $\phi(\cdot)$ is attached to the pre-trained encoder and mapping layer, and the entire network is fine-tuned. The trained model was evaluated on both the NTU RGB+D 60 and NTU RGB+D 120 datasets. The model was trained for 150 epochs with a batch size of 128 and an initial learning rate of 0.1. Table~\ref{tab6_2} compares the performance of DiMoP with the SOTA approaches. DiMoP achieved results comparable to those of the recent generative SOTA method. For instance, DiMoP increased accuracy from 97.3\% to 97.7\% in the NTU RGB+D 60 X-view evaluation. DiMoP achieved comparable results to MAMP~\cite{mao01} on the NTU RGB+D 120 dataset. Additionally, DiMoP outperformed the fully supervised STGCN~\cite{Yan01} and BlockGCN~\cite{blockgcn}  across all evaluations on the NTU RGB+D 60 dataset while achieving comparable results on the NTU RGB+D 120 dataset. 
	
	\begin{table}[thb]
		\caption[Supervised fine-tune evaluation of DiMoP ]{Supervised fine-tune evaluation results on the NTU RGB+D 60, and NTU RGB+D 120 datasets.}
		\label{tab6_2}
		\centering
		\begin{tabular}{@{}l |cc|cc}
			\cline{1-5}
			Models     & \multicolumn{2}{c}{NTU 60 (\%)} & \multicolumn{2}{|c}{ NTU 120 (\%)} \\

			&   X-sub &  X-view &  X-sub  &  X-set \\
			
			\cline{1-5} 
			
			\textit{Contrastive Methods} & & & & \\
			
			3s-CrosSCLR      & 86.2 & 92.5  & 80.5 & 80.4 \\
			3s-AimCLR        & 86.9 & 92.8  & 80.1   & 80.9\\
			3s-PSTL           & 87.1 & 93.9 & 81.3 & 82.6 \\
			3s-ActCLR     & 88.2  & 93.9 & 81.6   & 81.2\\
			3s-STJD-CL     &89.3 &94.8 &83.5 &86.8 \\
			\cline{1-5} 
			
			\textit{Generative Methods} & & & & \\
			SkeletonMAE & 88.5&94.7 &87.0 &88.9 \\
			MAMP & \underline{93.1}& 97.5&\underline{90.0} &\textbf{91.3 }\\
			MacDiff &92.7 &97.3 & -& -\\
			S-JEPA & \underline{93.1}&\underline{97.6} &\textbf{90.3} &\textbf{91.3} \\
			
			\cline{1-5} 
			\textcolor{gray}{\textit{Supervised Methods}} & & & & \\
			\textcolor{gray}{3s-STGCN~\cite{Yan01}}           & \textcolor{gray}{85.2} & \textcolor{gray}{91.4}  & \textcolor{gray}{77.2} & \textcolor{gray}{77.1} \\
			\textcolor{gray}{CTR-GCN~\cite{Chen2021a}} & \textcolor{gray}{92.4} & \textcolor{gray}{96.8} & \textcolor{gray}{88.9} & \textcolor{gray}{90.6} \\
			\textcolor{gray}{BlockGCN~\cite{blockgcn}} & \textcolor{gray}{93.1} & \textcolor{gray}{97.0} & \textcolor{gray}{90.3} & \textcolor{gray}{91.5} \\
			
			\cline{1-5} 
			\textbf{DiMoP (ours) }            & \textbf{93.2} & \textbf{97.7}   & \textbf{90.3} &\underline{91.1}\\
			\cline{1-5} 
		\end{tabular}
		
	\end{table}

	\subsubsection{Transfer Learning Evaluation}
	
	The effectiveness of the proposed method is evaluated in a transfer-learning setup to assess its generalization capability. The encoder and mapping layer are pre-trained on the NTU RGB+D 60 and NTU RGB+D 120 datasets separately using X-sub settings and are evaluated on the PKU-MMD II dataset using the supervised fine-tuning evaluation setting. The results are presented in Table~\ref{tab6_3}. DiMoP achieved state-of-the-art (SOTA) performance on NTU RGB+D 60, surpassing MacDiff by 0.2 percentage points. These improvements confirm that DiMoP generalizes well beyond the datasets and highlight its robustness under transfer learning.
	
	\begin{table} [thb]
		\caption[Transfer learning performance of the proposed DiMoP]{ Transfer learning evaluation of DiMoP on the PKU-MMD Part II dataset: Fine-tuning the pre-trained encoder from NTU RGB+D 60 and NTU RGB+D 120.}  
		\label{tab6_3}
		
		\centering
		\begin{tabular}{@{}l |cc}
			\cline{1-3} 
			Models    &   \multicolumn{2}{|c}{ To PKUMMD II (\%)} \\
			& NTU 60 &  NTU 120 \\
			\cline{1-3} 
			SkeletonMAE       & 58.4   & 61.0 \\
			MAMP & 70.6 & 73.2 \\
			MacDiff & \underline{72.2} & 73.4  \\
			S-JEPA & 71.4 & \textbf{74.2}  \\
			\cline{1-3} 
			\textbf{DiMoP (ours) }             &  \textbf{72.4}  &  \textbf{73.9} \\
			\cline{1-3} 
		\end{tabular}
		
	\end{table}

		
		

	\subsubsection{Semi-supervised evaluation}
	
	The post-attached classification layer and pre-trained encoder are fine-tuned using 1\% and 10\% of the training data, following the supervised settings in Section~\ref{supervse}. Table~\ref{6_4} compares our method with state-of-the-art approaches. The proposed DiMoP achieved a 2.6 percentage point improvement in accuracy over S-JEPA and yielded the same state-of-the-art (SOTA) result as S-JEPA on the NTU RGB+D 60 X-sub evaluation, utilizing only 1\% and 10\% of the training data, respectively. 
	

	
	Although MacDiff~\cite{MacDiff} reports higher semi-supervised results in its original paper, its exact semi-supervised training configuration was not provided with the released code. Hence, rather than directly comparing against potentially different settings, we re-trained MacDiff under the same semi-supervised protocol used for DiMoP, including identical label ratios, training settings, and evaluation protocols. The lower reproduced MacDiff accuracy in Table~\ref{6_4} than that of originally reported may be attributed to the different settings. Nevertheless, under this controlled comparison, DiMoP improves the reproduced MacDiff results by 0.3 and 0.4 percentage points on the X-view protocol with 1\% and 10\% labeled data, respectively.
	
	\begin{table}
		\caption{ Semi-supervised performance of DiMoP and its comparison with the SOTA methods.  \textit{$^*$ results were obtained by the provided codes}.}
		\label{6_4}
		\centering
		\begin{tabular}{l |cc|cc}
			
			\cline{1-5} 
			Models    & \multicolumn{4}{c}{NTU RGB+D 60 (\%)}  \\
			& \multicolumn{2}{c|}{\cellcolor{gray!20} \textit{1\% labels}} & \multicolumn{2}{c}{\cellcolor{gray!20} \textit{10\% labels}} \\
			
			&   X-sub &  X-view & X-sub & X-view \\
			
			\cline{1-5}

			3s-AimtCLR      & 54.8  & 54.3  & 78.2 & 81.6 \\
			3s-ActCLR  & 64.8 & 65.6  & 81.7 & 85.8 \\
			3s-STJD-CL & \underline{67.5} & 68.4 & 82.5& 88.0 \\
			SkeletonMAE & 54.4 &  54.6 & 80.6 & 83.5 \\
			MAMP & 66.0 & 68.7 & 88.0 & 91.5 \\
			MacDiff & 65.6 & \textbf{77.3} & \underline{88.2} & \textbf{92.5} \\
			\textit{MacDiff ${^*}$} & \textit{64.4} & \textit{70.9} & \textit{87.5} & \textit{91.9} \\
			S-JEPA & \underline{67.5} & 69.1 & \textbf{88.4} & 91.4 \\
			\cline{1-5} 
			\textbf{DiMoP (ours) }              & \textbf{70.1}  & \underline{71.2} & \textbf{88.4} & \underline{92.3} \\
			\cline{1-5} 
			
		\end{tabular}
		
	\end{table}

	\subsection{Ablation studies}
	\label{ablation6}
	The joint decoder configuration is used for the first four ablation studies.  Unless an ablation explicitly varied a factor (e.g., masking ratio, number of diffusion steps, decoder design, or loss weights), all other hyperparameters were fixed to the linear evaluation settings.
	
	\subsubsection{Analysis of masking strategy and masking ratio}
	Table~\ref{tab6_masking} compares simple random masking with motion-aware random masking, implemented following~\cite{mao01}. Random masking achieves better performance because it uniformly exposes all joints to the denoising objective, whereas motion-aware masking may over-emphasize large-motion joints and under-sample subtle yet discriminative motions. Nevertheless, uniform masking is not intended to explicitly model structured occlusion. In scenarios with persistent joint occlusion, such as lower-body occlusion behind a desk, joint-confidence scores from the pose estimator could be incorporated into both the masking probabilities and denoising loss to avoid treating unreliable or unavailable joints as reconstruction targets. Table~\ref{tab6_ratio} further shows that a high masking ratio of 90\% yields the best performance.
	

	\begin{table}[htbp]
		\centering
		\caption[DiMoP performance under varying masking strategies.]{DiMoP performance under varying masking strategies. The linear evaluation results are reported on the NTU RGB+D 60 Xsub dataset. The best one is in \textbf{bold}.}
		\label{tab6_masking}
		\begin{tabular}{l  | c }
			\hline
			Masking strategy & Acc(\%) \\
			\hline
			Random Masking  & \textbf{86.48} \\
			
			Motion Aware masking   & 86.13 \\
			
			\hline
		\end{tabular}
		
	\end{table}
	
	\begin{table}[htbp]
		\centering
		\caption[DiMoP performance under varying masking ratios.]{DiMoP performance under varying masking ratios. The linear evaluation results are reported on the NTU RGB+D 60 Xsub dataset. The best one is in \textbf{bold}.}
		\label{tab6_ratio}
		\begin{tabular}{l  | c }
			\hline
			Masking ratio & Acc(\%) \\
			\hline
			
			60\% & 70.21 \\
			65\% & 74.44 \\
			70\% & 75.65 \\
			75\% & 78.38 \\
			80\% & 82.59 \\
			85\% & 85.64 \\
			90\% & \textbf{86.48} \\
			95\% & 81.23 \\

			\hline
		\end{tabular}
		
	\end{table}
	
	\subsubsection{Hyperparameter tuning of the loss function and contribution analysis of each loss term}
	
	Table~\ref{tab6_hyper} reports the linear evaluation results under different loss-weight settings. The denoising loss is the primary generative objective, while the consistency and smoothness losses impose global and local constraints on the pseudo-classifier. We set $\alpha=\gamma=1$ to balance the generative and discriminative objectives. Since $\mathcal{L}_{smooth}$ is auxiliary to sequence-level consistency, $\lambda/\gamma$ should generally remain below 1. Our experiments showed that $\alpha=1$, $\lambda=0.3$, and $\gamma=1$ yielded the best performance across all datasets and evaluation protocols considered in this study. 
	
	As a practical guideline, we recommend fixing the other loss weights and setting $\lambda$ to a relatively small value for cases involving complex and diverse actions, while increasing them for cases involving less diverse actions with lower temporal variation.

	
	
	\begin{table}[htbp]
		\caption{Hyperparameter tuning for DiMoP on the NTU-RGB+D 60 dataset under the X-Sub setting in linear evaluation. The best one is in \textbf{bold}.}
		\label{tab6_hyper}
		\begin{center}
			\begin{tabular}{l l l c}
				\hline
				\textbf{$\alpha$} & $\lambda$ & $\gamma$ & \textbf{\textit{Acc \%}} \\
				\hline 
				
				1 &	1	&1&	85.85 \\
				1&	0.1	&0.1&	85.34 \\
				1&	1	&0.1&	85.81 \\
				1&	0.1	& 1	& 86.24 \\
				1&	0.3	& 1	& \textbf{86.48} \\
				1&	0.5	&0.5&	86.02 \\ 
				0.9&	0.1&	0.9&	86.35 \\
				0.5	&0.2	&0.2	&83.23 \\
				
				\hline
			\end{tabular}
		\end{center}

	\end{table}

	\begin{table}
		\centering
		\caption{Module Contribution Analysis in DiMoP}
		\label{tab6_component}

		\begin{tabular}{c  c  c |c}
			\hline
			$\mathcal{L}_{Denoise}$  & $\mathcal{L}_{smooth}$ & $\mathcal{L}_{consistency}$ & Acc (\%) \\
			\hline
			\checkmark &  & & 85.31 \\
			&  \checkmark & \checkmark &  34.73 \\
			\checkmark & \checkmark & & 85.81 \\
			\checkmark& & \checkmark & 86.09 \\
			\checkmark& \checkmark & \checkmark & \textbf{86.48 }\\
			
			\hline
		\end{tabular}
		
	\end{table}
	
	\subsubsection{Analysis of the number of diffusion steps and Noise scheduling}
	
\begin{figure}
		\centering
		\includegraphics[width=0.8\linewidth]{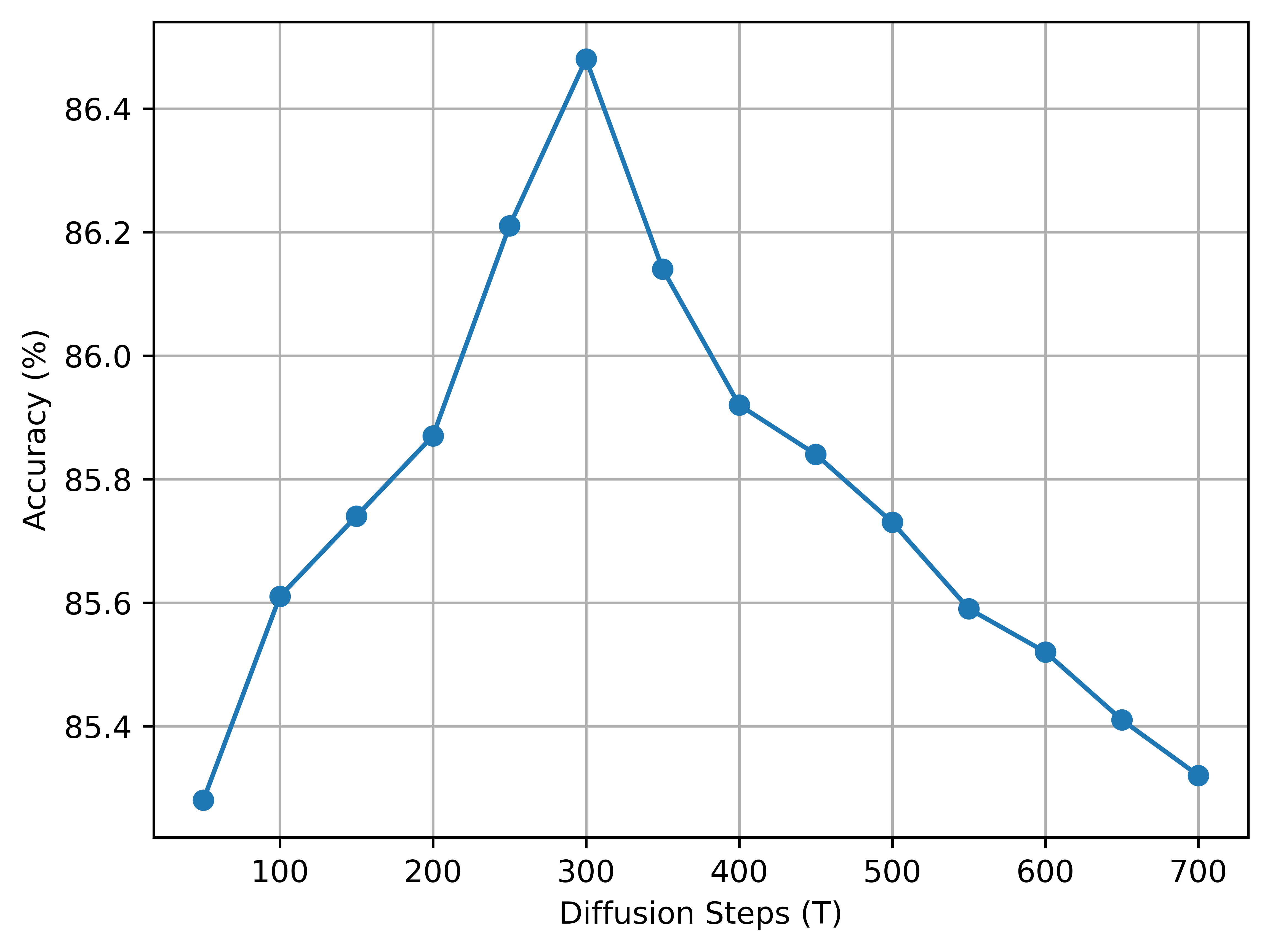}
		\caption[Ablation on diffusion steps.]{Linear evaluation performance variation against the number of diffusion steps on the NTU RGB+D X-sub.}
		\label{fig:diff_steps}
	\end{figure}

	The impact of noise level on learning discriminative features is studied by varying the number of diffusion steps $T$ from $T= 50$ to $T=700$ in 50-step increments. A linear schedule is used by default, following DDPM~\cite{ho01}, where the variances $\beta[1:T ]$ increase linearly. Figure ~\ref{fig:diff_steps} presents the linear evaluation accuracy on the NTU RGB+D X-sub evaluation with varying diffusion steps. It is observed that the model performed well with 250 to 350 diffusion steps compared to those with either low or high diffusion steps. When the number of diffusion steps is high, the diffused samples approach random noise, which prevents the denoising networks from learning meaningful and discriminative feature representations for downstream classification tasks.
	
	The selection of noise schedules, such as linear and cosine schedules~\cite{ho01}, for forward diffusion has been validated. Table~\ref{tab6_noise} presents the linear evaluation accuracy on NTU RGB+D X-sub evaluation with varying noise schedules. For this experiment, a noise level $T=300$ was used, and the joint decoder was employed. The linear schedule yields the best performance. 
	
	\begin{table}[htbp]
		\centering
		\caption[DiMoP performance under varying noise schedules. ]{DiMoP performance under varying noise schedules. The linear evaluation results are reported on the NTU RGB+D 60 Xsub dataset. The best one is in \textbf{bold}.}
		\label{tab6_noise}
		\begin{tabular}{l  | c }
			\hline
			Noise Schedule & Acc(\%) \\
			\hline
			Linear & \textbf{86.48} \\
			
			Cosine   &  85.61 \\
			
			\hline
		\end{tabular}
		
	\end{table}
	
	\subsubsection{Analysis of denoising prediction target:}
	
	In the denoising process, various prediction targets were evaluated, including noise, $\epsilon$ prediction, the prediction of the original joint coordinates, $x_0$, and the prediction of the motion dynamics of the original $x_0$. This work focuses on the impact of prediction targets on representation learning, rather than on the quality of generation. As demonstrated in Table~\ref{tab6_target}, predicting the motion dynamics of the original sample $x_0$ yielded better performance than noise prediction.

	\begin{table}[htbp]
		\centering
		\caption[Ablation study on the prediction target.]{Ablation study on the prediction target. The linear evaluation results are reported on the NTU RGB+D 60 Xsub dataset. The best one is in \textbf{bold}.}
		\label{tab6_target}
		\begin{tabular}{l  | c }
			\hline
			Prediction target & Acc(\%) \\
			\hline
			Noise $\epsilon$ & 49.03 \\
			Coordinates of the original $x_0$ & 85.94 \\ 
			Motion dynamics of the original $x_0$   & \textbf{86.48} \\
			
			\hline
		\end{tabular}
		
	\end{table}

	\subsubsection{Denoising decoder configurations}
	
	The three design configurations described in Section~\ref{6_decoder} were experimentally validated. During these experiments, $T=300$ was set, and a linear noise schedule was employed. The linear evaluation performance on the NTU RGB+D 60 X-Sub dataset is presented in Table~\ref{tab6_deco_design}. The best performance was obtained with Cross Decoder V2, thereby validating the theoretical design configuration.
	
	\begin{table}[htbp]
		\centering
		\caption[Ablation study on denoising decoder configurations.]{Comparison of Denoising Decoder Configurations: Linear Evaluation Results on the NTU RGB+D 60 Xsub Dataset with the Best Configuration in \textbf{Bold}.}
		\label{tab6_deco_design}
		\begin{tabular}{l  | c }
			\hline
			Configuration & Acc(\%) \\
			\hline
			Joint decoder  & 86.48 \\
			Cross decoder V1   & 85.73 \\
			Cross decoder V2   & \textbf{86.76}\\
			
			\hline
		\end{tabular}
		
	\end{table}


	
	\begin{figure*}[htbp]
		\centering
		\begin{subfigure}{0.3\textwidth}
			\includegraphics[width=\textwidth]{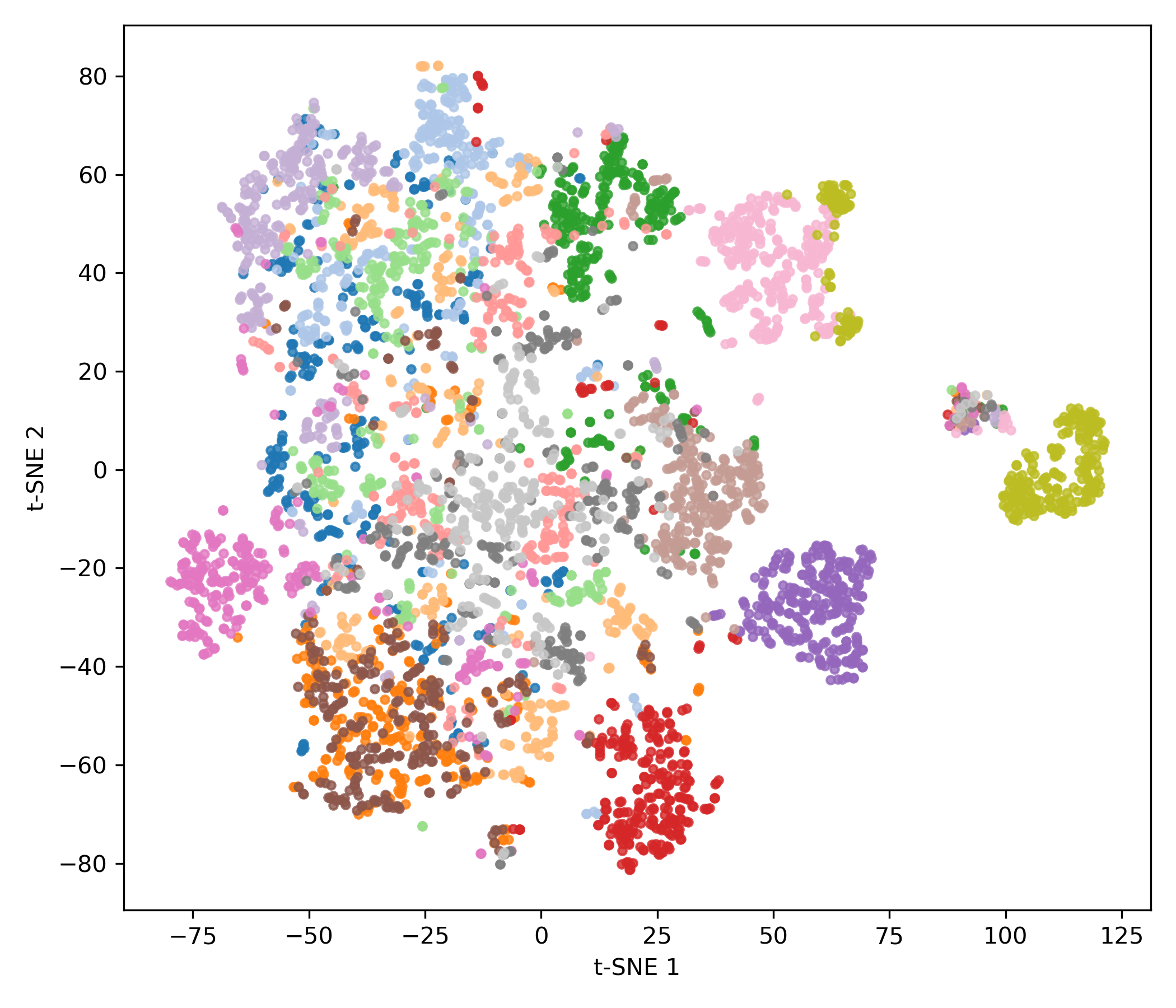}
			\caption{MAMP  }
			\label{figtsnr:sub1}
		\end{subfigure}
		\hfil
		\begin{subfigure}{0.3\textwidth}
			\includegraphics[width=\textwidth]{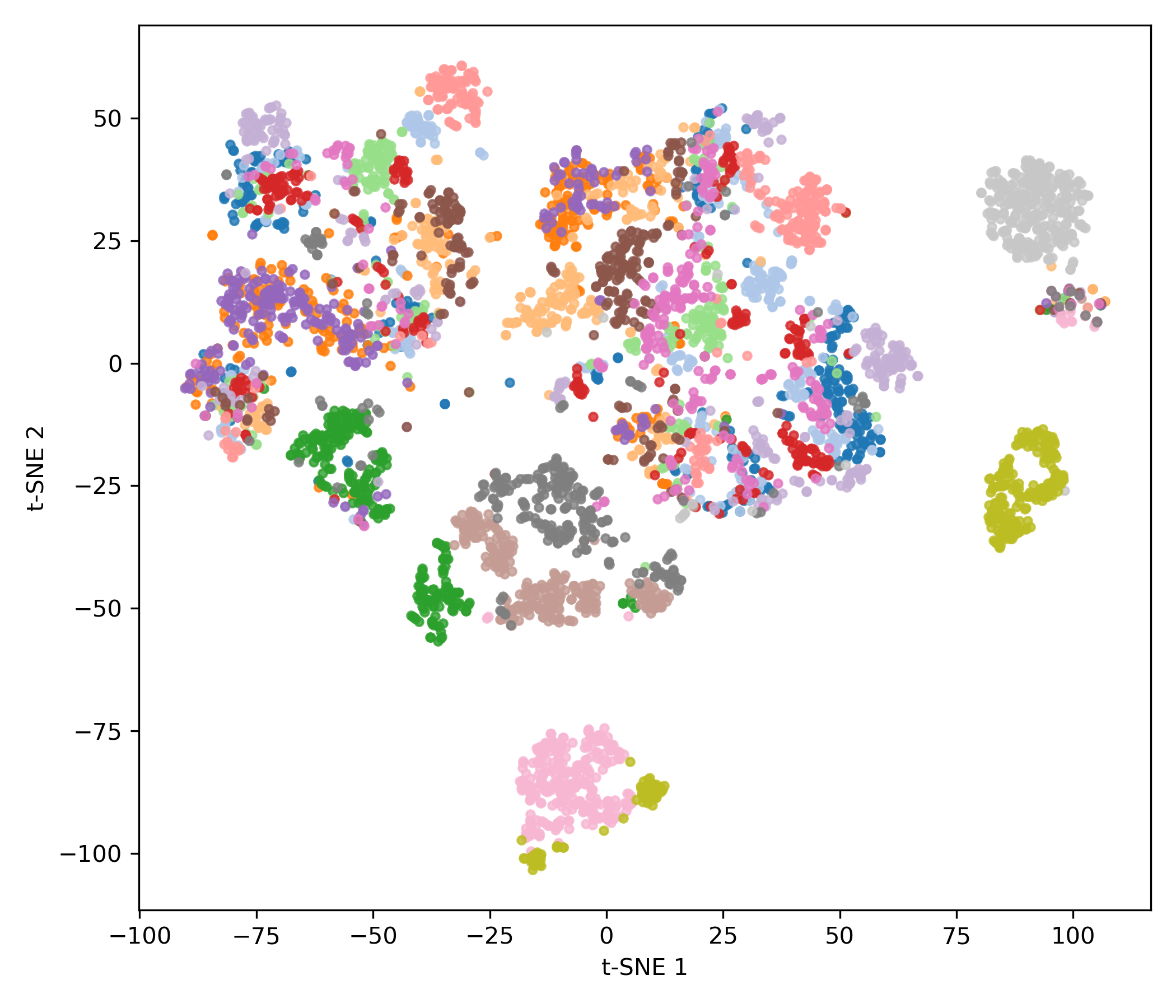}
			\caption{MacDiff}
			\label{figtsnr:sub2}
		\end{subfigure}
		\hfil
		\begin{subfigure}{0.3\textwidth}
			\includegraphics[width=\textwidth]{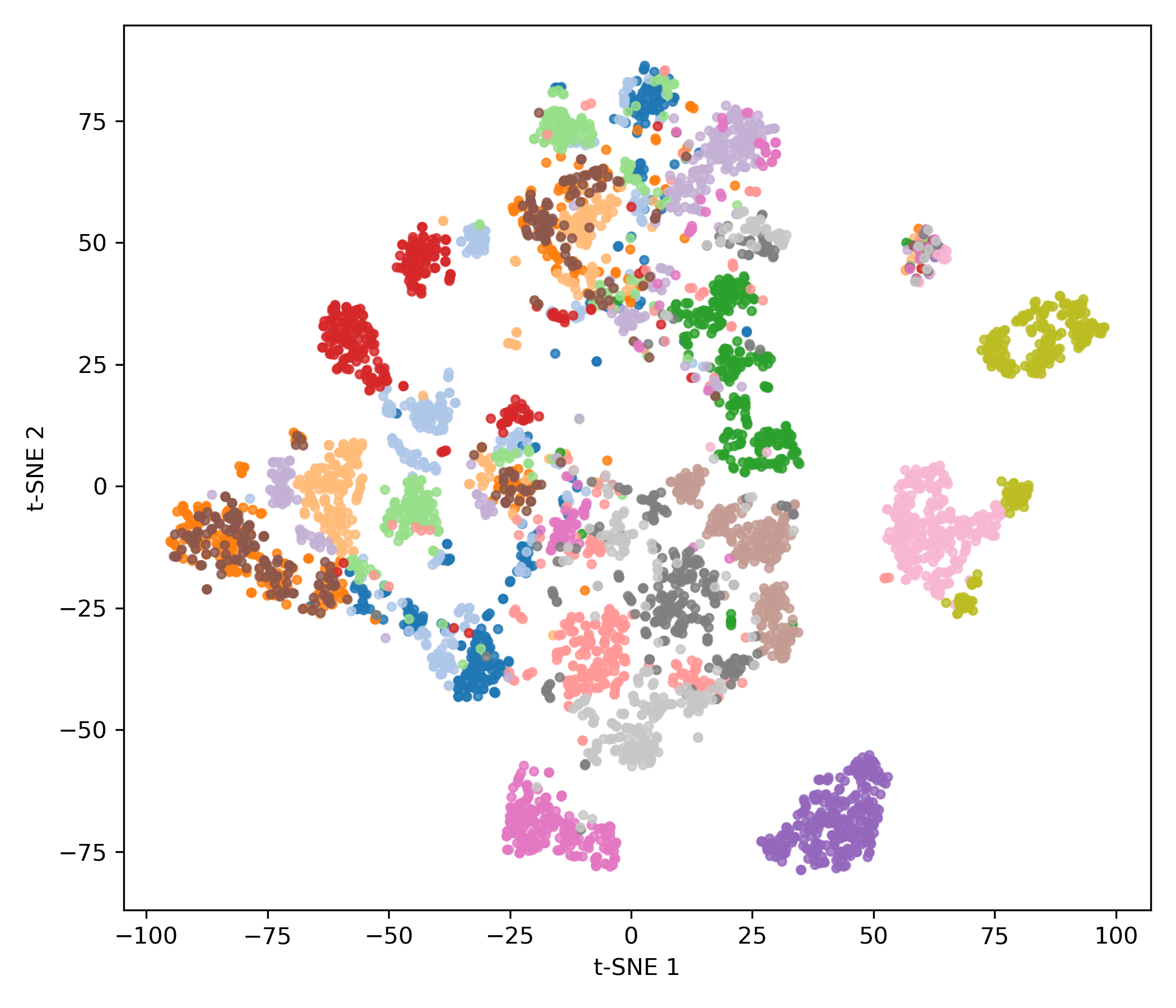}
			\caption{DiMoP }
			\label{figtsnr:sub3}
		\end{subfigure}

		
		{\footnotesize
			{\color{class2}$\blacksquare$} A2, 
			{\color{class3}$\blacksquare$} A3, 
			{\color{class11}$\blacksquare$} A11, 
			{\color{class12}$\blacksquare$} A12, 
			{\color{class14}$\blacksquare$} A14, 
			{\color{class16}$\blacksquare$} A16, 
			{\color{class17}$\blacksquare$} A17, 
			{\color{class29}$\blacksquare$} A29, 
			{\color{class30}$\blacksquare$} A30, 
			{\color{class31}$\blacksquare$} A31, 
			{\color{class33}$\blacksquare$} A33, 
			{\color{class34}$\blacksquare$} A34, 
			{\color{class36}$\blacksquare$} A36, 
			{\color{class41}$\blacksquare$} A41, 
			{\color{class44}$\blacksquare$} A44, 
			{\color{class45}$\blacksquare$} A45, 
			and {\color{class46}$\blacksquare$} A46.
		}
		
	
	\caption{The t-SNE visualization of embeddings on the NTU RGB+D 60 X-view benchmark. The same set of actions with subtle motion is used. (The MAMP ~\cite{mao01} and MacDiff~\cite{MacDiff} results are obtained by regenerating results with the provided code)}
	\label{fig:tsnr}
	
\end{figure*}

\subsubsection{Class-wise performance comparison on X-sub evaluation}

\begin{figure}
	\centering
	\includegraphics[scale= 0.4]{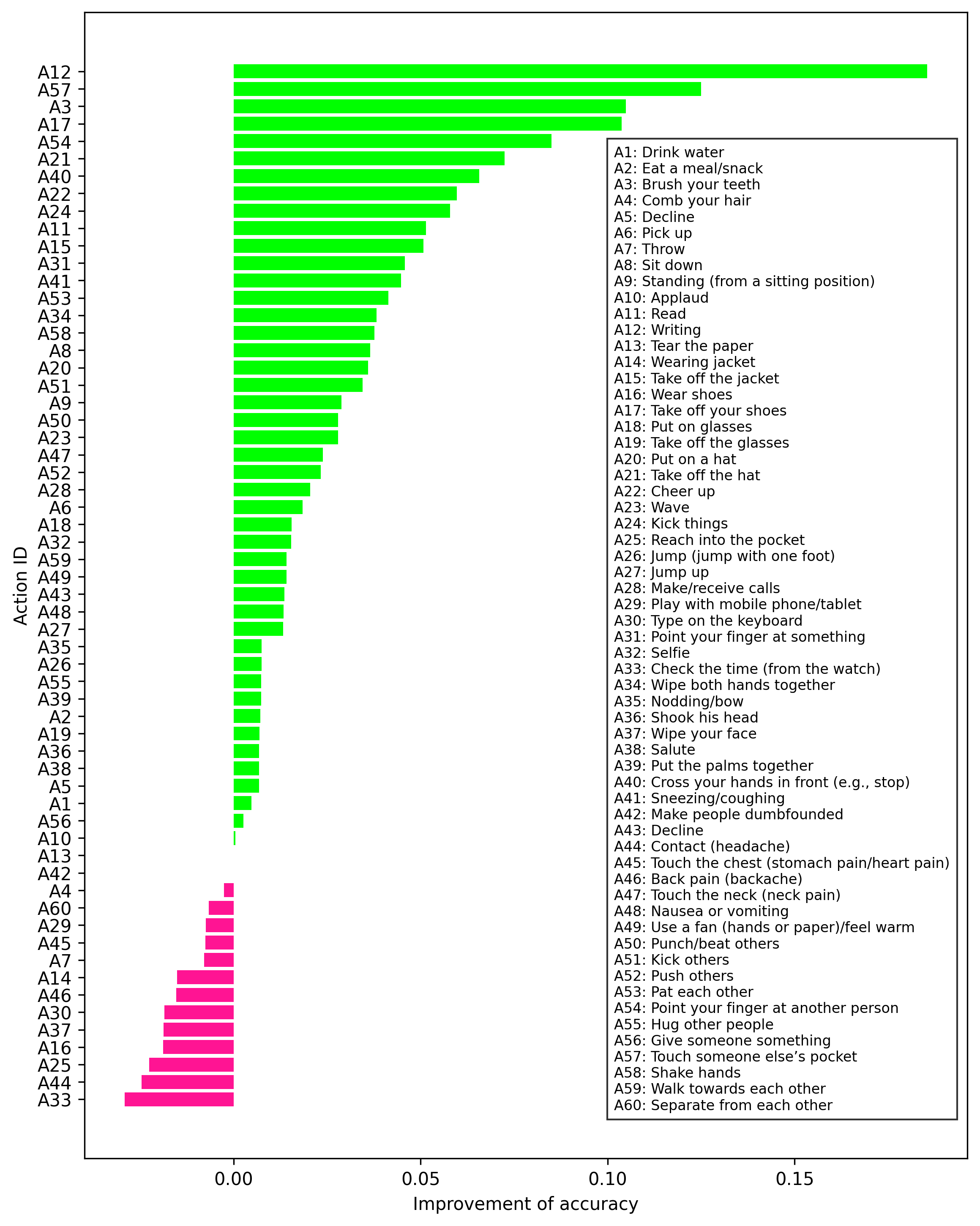}
	\caption[Class-wise accuracy difference on NTU RGB+D 60 X-sub]{Class-wise linear evaluation accuracy difference between DiMoP and MAMP on NTU RGB+D 60 X-sub.}
	\label{fig6_xsub_clsswise}
\end{figure}

Class-wise accuracies for the proposed DiMoP are compared with those of the baseline method, MAMP~\cite{mao01}, whose results were reproduced using publicly available code. On the NTU RGB+D 60 dataset, 47 of the 60 action classes show improved accuracy in the X-sub. Figure~\ref{fig6_xsub_clsswise} presents a detailed comparison, where pink indicates drops and green indicates gains. 

As shown, substantial improvements were observed for the majority of actions. Performance declines were noted for some actions, many of which involve interactions between two subjects. Actions in which the discriminative cue is subtle and short-lived, while most frames are dominated by neutral or overlapping poses, remain challenging. In these cases, sequence-level homogeneity tends to overshadow the brief but critical cues, resulting in misclassifications and reduced accuracy. These results underscore the effectiveness of DiMoP in modeling the distribution of joint dynamics.


\subsubsection{Performance Analysis Among Action Groups}
To provide a more fine-grained analysis, the 60 action classes are grouped according to their motion characteristics: fine-grained single-person actions (G1), large body-motion actions (G2), hand-object interactions (G3), two-person interactions (G4) and weak-discriminative-frame actions (G5). As summarized in Table~\ref{tab:groupwise_compact}, DiMoP improves 47 out of 60 classes, corresponding to 78.3\% of the actions, with an average gain of 1.90 percentage points over MAMP. Consistent improvements are observed across all groups, particularly for large body-motion actions and two-person interactions, where 91.7\% and 90.9\% of the classes are improved, respectively. These results indicate that progressive denoising facilitates the learning of both localized motion patterns and interaction-related dynamics. The relatively lower improvement ratios in hand-object and weak-discriminative-frame actions suggest that skeleton-only representations remain limited when object contact cues or short-lived discriminative frames are critical.

\begin{table}[t]
	\centering
	\caption{Group-wise comparison between DiMoP and MAMP on NTU RGB+D 60 X-sub under linear evaluation.}
	\label{tab:groupwise_compact}

	\begin{tabular}{lccc}
		\hline
		Group & Imp./Drop & \% imp & Mean $\Delta$Acc. \\
		\hline
		G1: Fine-grained single-person & 17/6 & 74\% & +1.88 \\
		G2: Large body-motion & 11/1& 92\% & +2.19 \\
		G3: Hand-object interaction & 9/5& 64\%& +2.59 \\
		G4: Two-person interaction & 10/1 & 91\%& +3.57 \\
		G5: Brief-discriminative-frame$^{\dagger}$ & 10/6& 63\% & +2.16 \\
		
		\hline
		Overall & 47/13 & 78\%& +1.90 \\
		\hline
	\end{tabular}
	
	\vspace{0.5mm}
		\footnotesize
		\textit{Action IDs:}
		\textbf{G1}: A3, A4, A18--A21, A28, A31, A33--A41, A44--A49;
		\textbf{G2}: A5--A9, A22--A24, A26, A27, A42, A43;
		\textbf{G3}: A1, A2, A10--A17, A25, A29, A30, A32;
		\textbf{G4}: A50--A60.
		\textbf{G5}: A3, A18--A21, A25, A28, A33, A37, A41, A44--A48, A57;
		
		$^{\dagger}$G5 overlaps with other groups and is used only for temporal failure analysis; the extended table is provided in the supplementary material.
\end{table}

	
	Figure~\ref{fig:tsnr} presents a t-SNE visualization of the learned embeddings for actions with subtle motion on the NTU RGB+D 60 dataset (X-Sub protocol). Compared with MAMP~\cite{mao01} and MacDiff~\cite{MacDiff}, DiMoP produces compact intra-class clusters and stronger inter-class separability. The embedding distribution shows that DiMoP captures the joint motion distributions more coherently, modeling subtle motion variations while suppressing cross-class overlap.

\section{Discussion and Conclusion}


A diffusion-based framework, DiMoP, was introduced for skeleton representation learning to model motion dynamics by progressively adding noise and denoising skeleton joints. By replacing static masking and handcrafted transformations with a stochastic, iterative denoising process, DiMoP effectively captures the uncertainty inherent in skeletal motion and enhances generalization. In addition, a pseudo-frame-based classifier was proposed to inject a discriminative structure without requiring manual labels. The proposed method achieved state-of-the-art performance on three standard benchmarks in action classification. 

However, DiMoP might be limited when discriminative cues are brief, highly localized, or dependent on object/contact information. For instance, actions such as \textit{headache} and \textit{neck pain} share similar hand-to-head/neck movements, where only a few frames carry discriminative evidence. The pseudo-frame classifier promotes sequence-level temporal consistency, which stabilizes representations but may suppress such short-lived cues. Moreover, object-centric actions such as \textit{wearing shoes}, \textit{typing on a keyboard}, or \textit{playing with a phone} may remain challenging because skeleton-only input does not explicitly encode object appearance or precise hand-object contact. It should be noted that these issues are not unique to DiMoP; they also exist in existing methods such as MAMP and MacDiff.

Future work will explore motion-adaptive diffusion scheduling and interaction between subjects. For instance, subtle moving joints will receive less noise or undergo slower denoising, while highly moving joints will retain full diffusion depth. This direction is expected to further improve the model's ability to distinguish between complex and diverse motion patterns in real-world scenarios.

			\bibliographystyle{unsrt}
			
			\bibliography{Bibiography}

@inproceedings{Chen2021a,
	title        = {Channel-wise Topology Refinement Graph Convolution for Skeleton-Based Action Recognition},
	author       = {Chen, Yuxin and Zhang, Ziqi and Yuan, Chunfeng and Li, Bing and Deng, Ying and Hu, Weiming},
	year         = 2021,
	booktitle    = {2021 IEEE/CVF International Conference on Computer Vision (ICCV)},
	volume       = {},
	number       = {},
	pages        = {13339--13348},
	doi          = {10.1109/ICCV48922.2021.01311}
}

@inproceedings{Yong0123,
	title        = {Hierarchical recurrent neural network for skeleton based action recognition},
	author       = {Yong Du and Wang, Wei and Wang, Liang},
	year         = 2015,
	booktitle    = {2015 IEEE Conference on Computer Vision and Pattern Recognition (CVPR)},
	volume       = {},
	number       = {},
	pages        = {1110--1118},
	doi          = {10.1109/CVPR.2015.7298714}
}

@inproceedings{Cheng2017,
	title        = {Skeleton-Based Action Recognition With Shift Graph Convolutional Network},
	author       = {Cheng, Ke and Zhang, Yifan and He, Xiangyu and Chen, Weihan and Cheng, Jian and Lu, Hanqing},
	year         = 2020,
	booktitle    = {2020 IEEE/CVF Conference on Computer Vision and Pattern Recognition (CVPR)},
	volume       = {},
	number       = {},
	pages        = {180--189},
	doi          = {10.1109/CVPR42600.2020.00026}
}

@inproceedings{Liu1905,
	title        = {NTU RGB+D 120: A Large-Scale Benchmark for 3D Human Activity Understanding},
	author       = {Liu, Jun and Shahroudy, Amir and Perez, Mauricio and Wang, Gang and Duan, Ling-Yu and Kot, Alex and Liu, • and Duan, L.-Y},
	year         = 2020,
	booktitle    = {{IEEE} Transactions on Pattern Analysis and Machine Intelligence},
	publisher    = {Institute of Electrical and Electronics Engineers ({IEEE})}
}

@inproceedings{Liu1703,
	title        = {PKU-MMD: A large scale benchmark for skeleton-based human action understanding},
	author       = {Liu, Chunhui and Hu, Yueyu and Li, Yanghao and Song, Sijie and Liu, Jiaying},
	year         = 2017,
	booktitle    = {Proceedings of the workshop on visual analysis in smart and connected communities},
	pages        = {1--8}
}

@article{wang01,
	title = {RGB-D-based human motion recognition with deep learning: A survey},
    journal = {Computer Vision and Image Understanding},
    volume = {171},
    pages = {118-139},
    year = {2018},
    issn = {1077-3142},
    doi = {https://doi.org/10.1016/j.cviu.2018.04.007},
    url = {https://www.sciencedirect.com/science/article/pii/S1077314218300663},
    author = {Pichao Wang and Wanqing Li and Philip Ogunbona and Jun Wan and Sergio Escalera}
    }

@article{wang03,
	title        = {Action recognition based on joint trajectory maps with convolutional neural networks},
	author       = {Pichao Wang and Wanqing Li and Chuankun Li and Yonghong Hou},
	year         = 2018,
	journal      = {Knowledge-Based Systems},
	volume       = 158,
	pages        = {43--53},
	doi          = {https://doi.org/10.1016/j.knosys.2018.05.029},
	issn         = {0950-7051},
	url          = {https://www.sciencedirect.com/science/article/pii/S0950705118302582}
}

@article{Chiara01,
	title        = {Skeleton-based action recognition via spatial and temporal transformer networks},
	author       = {Plizzari, Chiara and Cannici, Marco and Matteucci, Matteo},
	year         = 2021,
	journal      = {Computer Vision and Image Understanding},
	publisher    = {Elsevier},
	volume       = 208,
	pages        = 103219
}

@inproceedings{3mFormer,
	title        = {3Mformer: Multi-Order Multi-Mode Transformer for Skeletal Action Recognition},
	author       = {Wang, Lei and Koniusz, Piotr},
	year         = 2023,
	month        = {June},
	booktitle    = {Proceedings of the IEEE/CVF Conference on Computer Vision and Pattern Recognition (CVPR)},
	pages        = {5620--5631}
}

@inproceedings{shanaka01,
	title        = {Joint Temporal Pooling for Improving Skeleton-based Action Recognition},
	author       = {Gunasekara, Shanaka Ramesh and Li, Wanqing and Yang, Jack and  Ogunbona, Philip.},
	year         = 2023,
	booktitle    = {2023 International Conference on Digital Image Computing: Techniques and Applications (DICTA)},
	volume       = {},
	number       = {},
	pages        = {},
	doi          = {}
}

@article{Qiu01,
	title        = {Spatio-Temporal Tuples Transformer for Skeleton-Based Action Recognition},
	author       = {Helei Qiu and Biao Hou and Bo Ren and Xiaohua Zhang},
	year         = 2022,
	journal      = {ArXiv},
	volume       = {abs/2201.02849}
}

@article{attention,
	title        = {Attention is all you need},
	author       = {Vaswani, Ashish and Shazeer, Noam and Parmar, Niki and Uszkoreit, Jakob and Jones, Llion and Gomez, Aidan N and Kaiser, {\L}ukasz and Polosukhin, Illia},
	year         = 2017,
	journal      = {Advances in neural information processing systems},
	volume       = 30
}

@article{degcn,
	title        = {DeGCN: Deformable Graph Convolutional Networks for Skeleton-Based Action Recognition},
	author       = {Myung, Woomin and Su, Nan and Xue, Jing-Hao and Wang, Guijin},
	year         = 2024,
	journal      = {IEEE Transactions on Image Processing},
	volume       = 33,
	number       = {},
	pages        = {2477--2490},
	doi          = {10.1109/TIP.2024.3378886}
}

@inproceedings{blockgcn,
	title        = {BlockGCN: Redefining Topology Awareness for Skeleton-Based Action Recognition},
	author       = {Zhou, Yuxuan and Yan, Xudong and Cheng, Zhi-Qi and Yan, Yan and Dai, Qi and Hua, Xian-Sheng},
	year         = 2024,
	booktitle    = {Proceedings of the IEEE/CVF Conference on Computer Vision and Pattern Recognition}
}

@inproceedings{Yan01,
	title        = {Spatial Temporal Graph Convolutional Networks for Skeleton-Based Action Recognition},
	author       = {Yan, Sijie and Xiong, Yuanjun and Lin, Dahua},
	year         = 2018,
	booktitle    = {Proceedings of the Thirty-Second AAAI Conference on Artificial Intelligence and Thirtieth Innovative Applications of Artificial Intelligence Conference and Eighth AAAI Symposium on Educational Advances in Artificial Intelligence},
	location     = {New Orleans, Louisiana, USA},
	publisher    = {AAAI},
	series       = {AAAI'18/IAAI'18/EAAI'18},
	isbn         = {978-1-57735-800-8},
	articleno    = 912,
	numpages     = 9
}

@inproceedings{shahroudy01,
	title        = {Ntu rgb+ d: A large scale dataset for 3d human activity analysis},
	author       = {Shahroudy, Amir and Liu, Jun and Ng, Tian-Tsong and Wang, Gang},
	year         = 2016,
	booktitle    = {Proceedings of the IEEE conference on computer vision and pattern recognition},
	pages        = {1010--1019}
}

@inproceedings{aimclr,
  title={Contrastive learning from extremely augmented skeleton sequences for self-supervised action recognition},
  author={Guo, Tianyu and Liu, Hong and Chen, Zhan and Liu, Mengyuan and Wang, Tao and Ding, Runwei},
  booktitle={Proceedings of the AAAI conference on artificial intelligence},
  volume={36},
  number={1},
  pages={762--770},
  year={2022}
}

@inproceedings{skelemixclr,
	title        = {Contrastive Learning from Spatio-Temporal Mixed Skeleton Sequences for Self-Supervised Skeleton-Based Action Recognition},
	author       = {Zhan, Chen and Hong, Liu and Tianyu, Guo and Zhengyan, Chen and Pinhao, Song and Hao, Tang},
	year         = 2022,
	booktitle    = {arXiv}
}

@INPROCEEDINGS {actionlet,
author = { Lin, Lilang and Zhang, Jiahang and Liu, Jiaying },
booktitle = { 2023 IEEE/CVF Conference on Computer Vision and Pattern Recognition (CVPR) },
title = {{ Actionlet-Dependent Contrastive Learning for Unsupervised Skeleton-Based Action Recognition }},
year = {2023},
volume = {},
ISSN = {},
pages = {2363-2372},
doi = {10.1109/CVPR52729.2023.00234},
url = {https://doi.ieeecomputersociety.org/10.1109/CVPR52729.2023.00234},
publisher = {IEEE Computer Society},
address = {Los Alamitos, CA, USA},
month =Jun}

@inproceedings{crossclr,
  title={3d human action representation learning via cross-view consistency pursuit},
  author={Li, Linguo and Wang, Minsi and Ni, Bingbing and Wang, Hang and Yang, Jiancheng and Zhang, Wenjun},
  booktitle={Proceedings of the IEEE/CVF conference on computer vision and pattern recognition},
  pages={4741--4750},
  year={2021}
}

@inproceedings{thoker01,
	title        = {Skeleton-contrastive 3D action representation learning},
	author       = {Thoker, Fida Mohammad and Doughty, Hazel and Snoek, Cees GM},
	year         = 2021,
	booktitle    = {Proceedings of the 29th ACM international conference on multimedia},
	pages        = {1655--1663}
}

@inproceedings{pstl,
	title        = {Self-supervised action representation learning from partial spatio-temporal skeleton sequences},
	author       = {Zhou, Yujie and Duan, Haodong and Rao, Anyi and Su, Bing and Wang, Jiaqi},
	year         = 2023,
	booktitle    = {Proceedings of the Thirty-Seventh AAAI Conference on Artificial Intelligence and Thirty-Fifth Conference on Innovative Applications of Artificial Intelligence and Thirteenth Symposium on Educational Advances in Artificial Intelligence},
	publisher    = {AAAI Press},
	series       = {AAAI'23/IAAI'23/EAAI'23},
	doi          = {10.1609/aaai.v37i3.25495},
	isbn         = {978-1-57735-880-0},
	url          = {https://doi.org/10.1609/aaai.v37i3.25495},
	articleno    = 426,
	numpages     = 9
}

@inproceedings{glformer,
	title        = {Global-Local Motion Transformer for Unsupervised Skeleton-Based Action Learning},
	author       = {Kim, Boeun and Chang, Hyung Jin and Kim, Jungho and Choi, Jin Young},
	year         = 2022,
	booktitle    = {Computer Vision – ECCV 2022: 17th European Conference, Tel Aviv, Israel, October 23–27, 2022, Proceedings, Part IV},
	location     = {Tel Aviv, Israel},
	publisher    = {Springer-Verlag},
	address      = {Berlin, Heidelberg},
	pages        = {209–225},
	doi          = {10.1007/978-3-031-19772-7_13},
	isbn         = {978-3-031-19771-0},
	url          = {https://doi.org/10.1007/978-3-031-19772-7_13},
	numpages     = 17
}

@inproceedings{cpm,
	title        = {Contrastive Positive Mining for Unsupervised 3D Action Representation Learning},
	author       = {Zhang, Haoyuan and Hou, Yonghong and Zhang, Wenjing and Li, Wanqing},
	year         = 2022,
	booktitle    = {Computer Vision -- ECCV 2022},
	publisher    = {Springer Nature Switzerland},
	address      = {Cham},
	pages        = {36--51},
	isbn         = {978-3-031-19772-7},
	editor       = {Avidan, Shai and Brostow, Gabriel and Ciss{\'e}, Moustapha and Farinella, Giovanni Maria and Hassner, Tal}
}

@inproceedings{mao01,
  title={Masked Motion Predictors are Strong 3D Action Representation Learners},
  author={Mao, Yunyao and Deng, Jiajun and Zhou, Wengang and Fang, Yao and Ouyang, Wanli and Li, Houqiang},
  booktitle={2023 IEEE/CVF International Conference on Computer Vision (ICCV)},
  pages={10147--10157},
  year={2023},
  organization={IEEE}
}

@inproceedings{SkeletonMAE,
	title        = {SkeletonMAE: Graph-based Masked Autoencoder for Skeleton Sequence Pre-training},
	author       = {Yan, Hong and Liu, Yang and Wei, Yushen and Li, Zhen and Li, Guanbin and Lin, Liang},
	year         = 2023,
	month        = {October},
	booktitle    = {Proceedings of the IEEE/CVF International Conference on Computer Vision (ICCV)},
	pages        = {5606--5618}
}

@inproceedings{Wenhan01,
	title        = {Skeletonmae: Spatial-Temporal Masked Autoencoders for Self-Supervised Skeleton Action Recognition},
	author       = {Wu, Wenhan and Hua, Yilei and Zheng, Ce and Wu, Shiqian and Chen, Chen and Lu, Aidong},
	year         = 2023,
	booktitle    = {2023 IEEE International Conference on Multimedia and Expo Workshops (ICMEW)},
	volume       = {},
	number       = {},
	pages        = {224--229},
	doi          = {10.1109/ICMEW59549.2023.00045}
}

@inproceedings{ho01,
	title        = {Denoising Diffusion Probabilistic Models},
	author       = {Ho, Jonathan and Jain, Ajay and Abbeel, Pieter},
	year         = 2020,
	booktitle    = {Advances in Neural Information Processing Systems},
	publisher    = {Curran Associates, Inc.},
	volume       = 33,
	pages        = {6840--6851},
	url          = {https://proceedings.neurips.cc/paper_files/paper/2020/file/4c5bcfec8584af0d967f1ab10179ca4b-Paper.pdf},
	editor       = {H. Larochelle and M. Ranzato and R. Hadsell and M.F. Balcan and H. Lin}
}

@inproceedings{Jascha01,
	title        = {Deep unsupervised learning using nonequilibrium thermodynamics},
	author       = {Sohl-Dickstein, Jascha and Weiss, Eric A. and Maheswaranathan, Niru and Ganguli, Surya},
	year         = 2015,
	booktitle    = {Proceedings of the 32nd International Conference on International Conference on Machine Learning - Volume 37},
	location     = {Lille, France},
	publisher    = {JMLR.org},
	series       = {ICML'15},
	pages        = {2256–2265},
	numpages     = 10
}

@inproceedings{motionbert,
	title        = {MotionBERT: A Unified Perspective on Learning Human Motion Representations},
	author       = {Zhu, Wentao and Ma, Xiaoxuan and Liu, Zhaoyang and Liu, Libin and Wu, Wayne and Wang, Yizhou},
	year         = 2023,
	month        = {October},
	booktitle    = {Proceedings of the IEEE/CVF International Conference on Computer Vision (ICCV)},
	pages        = {15085--15099}
}

@article{hou01,
	title        = {Hypergraph transformer for skeleton-based action recognition},
	author       = {Zhou, Yuxuan and Cheng, Zhi-Qi and Li, Chao and Fang, Yanwen and Geng, Yifeng and Xie, Xuansong and Keuper, Margret},
	year         = 2022,
	journal      = {arXiv preprint arXiv:2211.09590}
}

@InProceedings{macdiff,
author="Wu, Lehong
and Lin, Lilang
and Zhang, Jiahang
and Ma, Yiyang
and Liu, Jiaying",
editor="Leonardis, Ale{\v{s}}
and Ricci, Elisa
and Roth, Stefan
and Russakovsky, Olga
and Sattler, Torsten
and Varol, G{\"u}l",
title="MacDiff: Unified Skeleton Modeling with Masked Conditional Diffusion",
booktitle="Computer Vision -- ECCV 2024",
year="2025",
publisher="Springer Nature Switzerland",
address="Cham",
pages="110--128",
isbn="978-3-031-73347-5"
}

@inproceedings{mae,
	title        = {Masked Autoencoders Are Scalable Vision Learners},
	author       = {He, Kaiming and Chen, Xinlei and Xie, Saining and Li, Yanghao and Doll\'ar, Piotr and Girshick, Ross},
	year         = 2022,
	month        = {June},
	booktitle    = {Proceedings of the IEEE/CVF Conference on Computer Vision and Pattern Recognition (CVPR)},
	pages        = {16000--16009}
}

@inproceedings{Vit,
	author       = {Alexey Dosovitskiy and
                  Lucas Beyer and
                  Alexander Kolesnikov and
                  Dirk Weissenborn and
                  Xiaohua Zhai and
                  Thomas Unterthiner and
                  Mostafa Dehghani and
                  Matthias Minderer and
                  Georg Heigold and
                  Sylvain Gelly and
                  Jakob Uszkoreit and
                  Neil Houlsby},
  title        = {An Image is Worth 16x16 Words: Transformers for Image Recognition
                  at Scale},
  booktitle    = {9th International Conference on Learning Representations, {ICLR} 2021,
                  Virtual Event, Austria, May 3-7, 2021},
  publisher    = {OpenReview.net},
  year         = {2021},
  url          = {https://openreview.net/forum?id=YicbFdNTTy},
  bibsource    = {dblp computer science bibliography, https://dblp.org}
}

@article{jtg,
	title        = {Spatio-Temporal Fusion for Human Action Recognition via Joint Trajectory Graph},
	author       = {Zheng, Yaolin and Huang, Hongbo and Wang, Xiuying and Yan, Xiaoxu and Xu, Longfei},
	year         = 2024,
	month        = {Mar.},
	journal      = {Proceedings of the AAAI Conference on Artificial Intelligence},
	volume       = 38,
	number       = 7,
	pages        = {7579--7587},
	doi          = {10.1609/aaai.v38i7.28590},
	url          = {https://ojs.aaai.org/index.php/AAAI/article/view/28590},
	abstractnote = {Graph Convolutional Networks (GCNs) and Transformers have been widely applied to skeleton-based human action recognition, with each offering unique advantages in capturing spatial relationships and long-range dependencies. However, for most GCN methods, the construction of topological structures relies solely on the spatial information of human joints, limiting their ability to directly capture richer spatio-temporal dependencies. Additionally, the self-attention modules of many Transformer methods lack topological structure information, restricting the robustness and generalization of the models. To address these issues, we propose a Joint Trajectory Graph (JTG) that integrates spatio-temporal information into a uniform graph structure. We also present a Joint Trajectory GraphFormer (JT-GraphFormer), which directly captures the spatio-temporal relationships among all joint trajectories for human action recognition. To better integrate topological information into spatio-temporal relationships, we introduce a Spatio-Temporal Dijkstra Attention (STDA) mechanism to calculate relationship scores for all the joints in JTG. Furthermore, we incorporate the Koopman operator into the classification stage to enhance the model’s representation ability and classification performance. Experiments demonstrate that JT-GraphFormer achieves outstanding performance in human action recognition tasks, outperforming state-of-the-art methods on the NTU RGB+D, NTU RGB+D 120, and N-UCLA datasets.}
}

@inproceedings{skateformer,
	 title={Skateformer: skeletal-temporal transformer for human action recognition},
      author={Do, Jeonghyeok and Kim, Munchurl},
      booktitle={European Conference on Computer Vision},
      pages={401--420},
      year={2025},
      organization={Springer}
}

@article{shanaka02,
	title        = {Asynchronous Joint-based Temporal Pooling for Skeleton-based Action Recognition},
	author       = {Gunasekara, Shanaka Ramesh and Li, Wanqing and Yang, Jack and Ogunbona, Philip},
	year         = 2024,
	journal      = {IEEE Transactions on Circuits and Systems for Video Technology},
	volume       = {},
	number       = {},
	pages        = {1--1},
	doi          = {10.1109/TCSVT.2024.3465845}
}

@inproceedings{gap,
	title={Generative action description prompts for skeleton-based action recognition},
  author={Xiang, Wangmeng and Li, Chao and Zhou, Yuxuan and Wang, Biao and Zhang, Lei},
  booktitle={Proceedings of the IEEE/CVF International Conference on Computer Vision},
  pages={10276--10285},
  year={2023}
}

@ARTICLE{Bruno,
  author={Degardin, Bruno and Lopes, Vasco and Proença, Hugo},
  journal={IEEE Transactions on Biometrics, Behavior, and Identity Science}, 
  title={Fake It Till You Recognize It: Quality Assessment for Human Action Generative Models}, 
  year={2024},
  volume={6},
  number={2},
  pages={261-271},
  doi={10.1109/TBIOM.2024.3375453}}

@ARTICLE{Bulat,
  author={Khaertdinov, Bulat and Asteriadis, Stylianos and Ghaleb, Esam},
  journal={IEEE Transactions on Biometrics, Behavior, and Identity Science}, 
  title={Dynamic Temperature Scaling in Contrastive Self-Supervised Learning for Sensor-Based Human Activity Recognition}, 
  year={2022},
  volume={4},
  number={4},
  pages={498-507},
  doi={10.1109/TBIOM.2022.3180591}}

@inproceedings{Wang05,
  title={Action recognition based on joint trajectory maps using convolutional neural networks},
  author={Wang, Pichao and Li, Zhaoyang and Hou, Yonghong and Li, Wanqing},
  booktitle={Proceedings of the 24th ACM international conference on Multimedia},
  pages={102--106},
  year={2016}
}

@inproceedings{NUCLA,
  title     = {Cross-View Action Modeling, Learning, and Recognition},
  author    = {Wang, Jingyu and Nie, Yin and Xia, Tianyang and Wu, Ying and Zhu, Song-Chun},
  booktitle = {Proceedings of the IEEE Conference on Computer Vision and Pattern Recognition (CVPR)},
  pages     = {2649--2656},
  year      = {2014},
  doi       = {10.1109/CVPR.2014.336}
}

@INPROCEEDINGS{CNN1,
  author={Du, Yong and Fu, Yun and Wang, Liang},
  booktitle={2015 3rd IAPR Asian Conference on Pattern Recognition (ACPR)}, 
  title={Skeleton based action recognition with convolutional neural network}, 
  year={2015},
  volume={},
  number={},
  pages={579-583},
  doi={10.1109/ACPR.2015.7486569}}

@ARTICLE{LSTM01,
  author={Du, Yong and Fu, Yun and Wang, Liang},
  journal={IEEE Transactions on Image Processing}, 
  title={Representation Learning of Temporal Dynamics for Skeleton-Based Action Recognition}, 
  year={2016},
  volume={25},
  number={7},
  pages={3010-3022},
  doi={10.1109/TIP.2016.2552404}}

@ARTICLE{chuankun02,
  author={Li, Chuankun and Li, Shuai and Gao, Yanbo and Gao, Xingyu and Chen, Ping and Li, Jian and Li, Wanqing},
  journal={IEEE Transactions on Circuits and Systems for Video Technology}, 
  title={Unsupervised Feature Enrichment and Fidelity Preservation Learning Framework for Skeleton-based Action Recognition}, 
  year={2025},
  volume={},
  number={},
  pages={1-1},
  doi={10.1109/TCSVT.2025.3540292}}

@inproceedings{USDRL,
  title={USDRL: Unified Skeleton-Based Dense Representation Learning with Multi-Grained Feature Decorrelation},
  author={Wanjiang Weng and Hongsong Wang and Junbo Wang and Lei He and Guosen Xie},
  booktitle={Proceedings of the AAAI Conference on Artificial Intelligence},
  year={2025}
}

@INPROCEEDINGS{TAHAR,
  author={Lerch, David J. and Zhong, Zeyun and Martin, Manuel and Voit, Michael and Beyerer, Jürgen},
  booktitle={2024 IEEE/CVF Winter Conference on Applications of Computer Vision Workshops (WACVW)}, 
  title={Unsupervised 3D Skeleton-Based Action Recognition using Cross-Attention with Conditioned Generation Capabilities}, 
  year={2024},
  volume={},
  number={},
  pages={202-211},
  doi={10.1109/WACVW60836.2024.00027}}

@ARTICLE{sltnNet,
  author={Jiang, Shengqin and Zhang, Haokui and Qi, Yuankai and Liu, Qingshan},
  journal={IEEE Transactions on Industrial Informatics}, 
  title={Spatial-Temporal Interleaved Network for Efficient Action Recognition}, 
  year={2025},
  volume={21},
  number={1},
  pages={178-187},
  doi={10.1109/TII.2024.3450021}}

@ARTICLE{d3d,
  author={Jiang, Shengqin and Qi, Yuankai and Zhang, Haokui and Bai, Zongwen and Lu, Xiaobo and Wang, Peng},
  journal={IEEE Transactions on Industrial Informatics}, 
  title={D3D: Dual 3-D Convolutional Network for Real-Time Action Recognition}, 
  year={2021},
  volume={17},
  number={7},
  pages={4584-4593},
  doi={10.1109/TII.2020.3018487}}

@ARTICLE{shanaka03,
  author={Gunasekara, Shanaka Ramesh and Li, Wanqing and Ogunbona, Philip and Yang, Jack},
  journal={IEEE Transactions on Biometrics, Behavior, and Identity Science}, 
  title={Spatio-Temporal Joint Density Driven Learning for Skeleton-Based Action Recognition}, 
  year={2025},
  volume={},
  number={},
  pages={1-1},
  doi={10.1109/TBIOM.2025.3566212}}

@article{LiKBS01,
title = {Variation-aware directed graph convolutional networks for skeleton-based action recognition},
journal = {Knowledge-Based Systems},
volume = {302},
pages = {112319},
year = {2024},
issn = {0950-7051},
doi = {https://doi.org/10.1016/j.knosys.2024.112319},
url = {https://www.sciencedirect.com/science/article/pii/S0950705124009535},
author = {Tianchen Li and Pei Geng and Guohui Cai and Xinran Hou and Xuequan Lu and Lei Lyu}
}

@article{LiKBS02,
title = {Skeleton-based action recognition through attention guided heterogeneous graph neural network},
journal = {Knowledge-Based Systems},
volume = {309},
pages = {112868},
year = {2025},
issn = {0950-7051},
doi = {https://doi.org/10.1016/j.knosys.2024.112868},
url = {https://www.sciencedirect.com/science/article/pii/S0950705124015028},
author = {Tianchen Li and Pei Geng and Xuequan Lu and Wanqing Li and Lei Lyu}
}

@article{KilicKBS01,
title = {AGMS-GCN: Attention-guided multi-scale graph convolutional networks for skeleton-based action recognition},
journal = {Knowledge-Based Systems},
volume = {311},
pages = {113045},
year = {2025},
issn = {0950-7051},
doi = {https://doi.org/10.1016/j.knosys.2025.113045},
url = {https://www.sciencedirect.com/science/article/pii/S0950705125000929},
author = {Ugur Kilic and Ozge Oztimur Karadag and Gulsah Tumuklu Ozyer}
}

			%
			
			\vskip 0pt plus -1fil
			\vspace{-6mm}
			\begin{IEEEbiography}[{\includegraphics[width=1in,height=1.25in,clip,keepaspectratio]{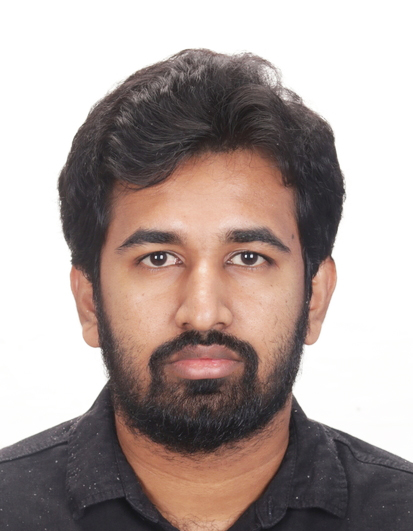}}]{Shanaka Ramesh Gunasekara}  (Member, IEEE) received his PhD in computer science and engineering with the Advanced Multimedia Research Lab (AMRL), University of Wollongong, Australia, with a focus on human action recognition. He received the B.Sc. (hons) degree in electrical and electronic engineering from the University of Peradeniya, Sri Lanka. He is currently working as a postdoc research fellow at RMIT. His research interests include 3D computer vision, human motion analysis, signal processing, medical image analysis, and robotics.\end{IEEEbiography}
			\vskip 0pt plus -1fil
			\vspace{-8mm}
			\begin{IEEEbiography}[{\includegraphics[width=1in,height=1.25in,clip,keepaspectratio]{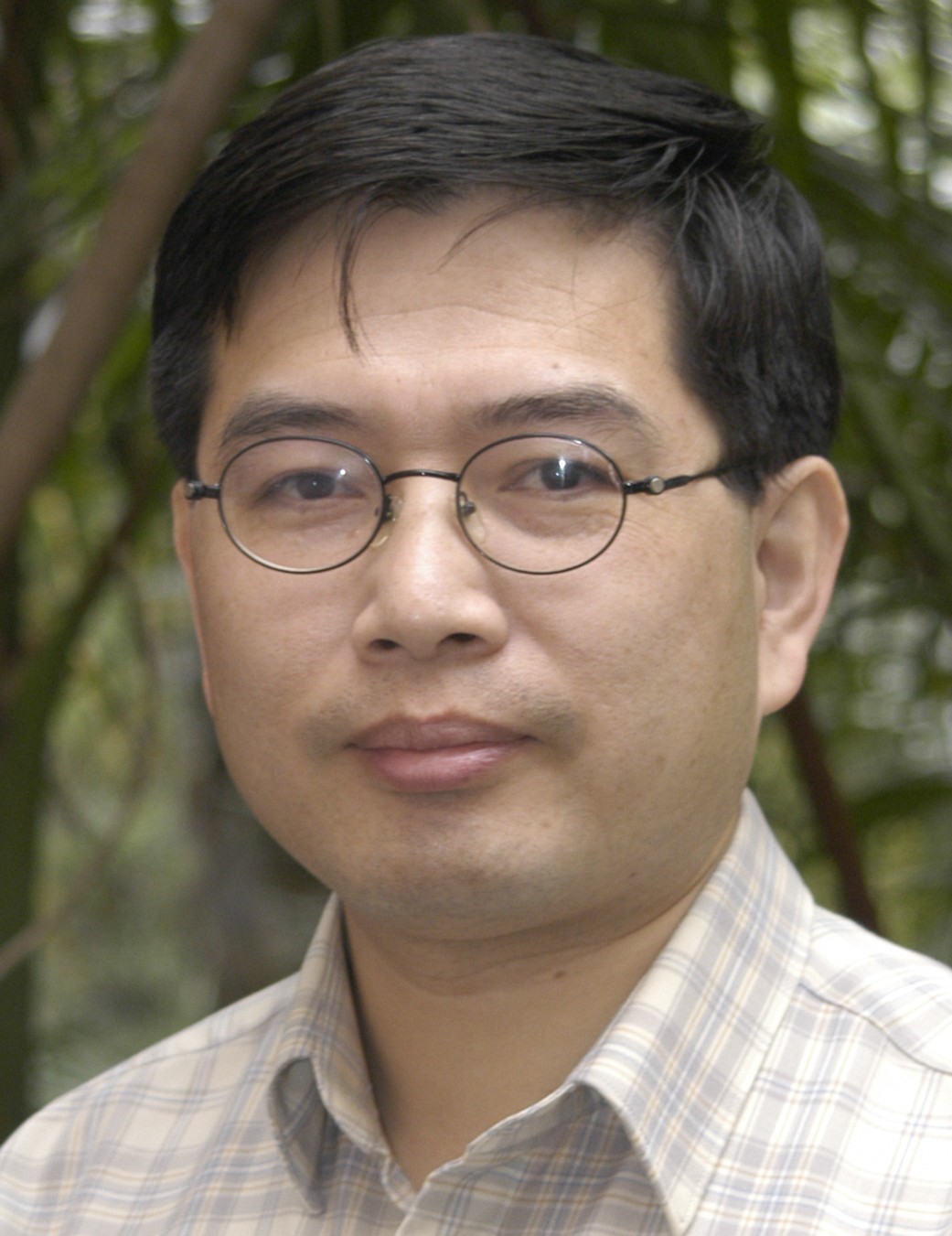}}]{Wanqing Li} (M’97-SM'05) received his PhD in electronic engineering from the University of Western Australia. He was a Senior Researcher and later a Principal Researcher at the Motorola Research Lab in Sydney from 1998 to 2003, and a visiting researcher at Microsoft Research, USA, in 2008, 2010, and 2013. He is currently a Professor and Co-Director of the Advanced Multimedia Research Lab (AMRL), University of Wollongong, Australia. His research areas include machine learning, 3D computer vision, 3D multimedia signal processing, medical image analysis, natural language processing, and their applications.
				Dr. Li served as a Technical Program Co-Chair for IEEE ICME 2021 and has served as Co-Chair for many IEEE Workshops. He is an Associate Editor for IEEE Transactions on Image Processing and IEEE Transactions on Multimedia. He served as an Associate Editor for IEEE Transactions on Circuits and Systems for Video Technology from 2018 to 2021 and for the Journal of Visual Communication and Image Representation from 2016 to 2019.\end{IEEEbiography}
			\vskip 0pt plus -1fil
			\vspace{-8mm}
			\begin{IEEEbiography}[{\includegraphics[width=1in,height=1.25in,clip,keepaspectratio]{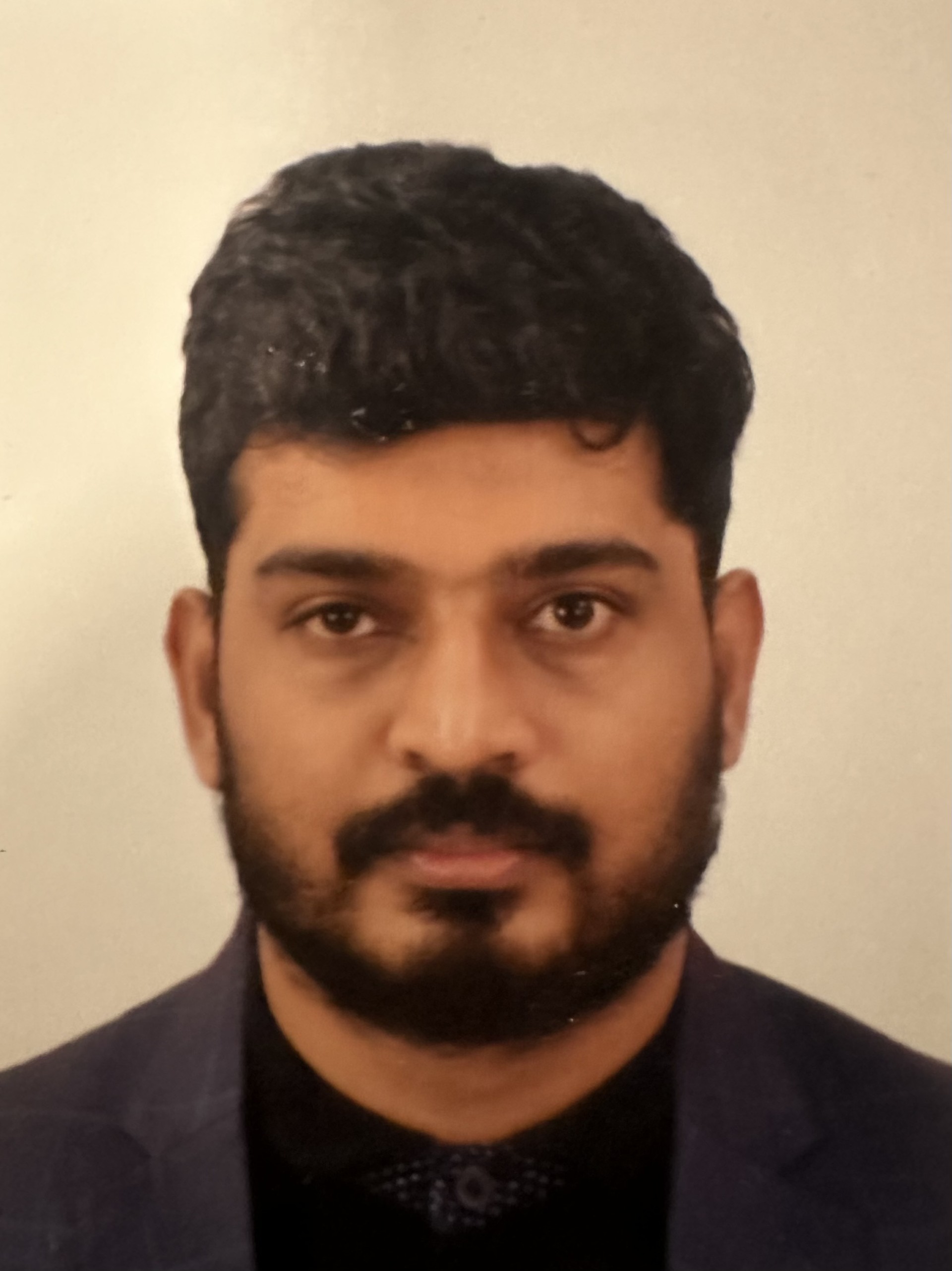}}]{Nikalal Kaldera} holds a Bachelor's degree in Electrical and Electronic Engineering from the University of Peradeniya, Sri Lanka. He is currently pursuing a Ph.D. at the Advanced Multimedia Research Laboratory (AMRL), University of Wollongong, Australia, focusing on human action recognition. His research interests include computer vision, machine learning, data science, and signal/image processing..\end{IEEEbiography}
			\vskip 0pt plus -1fil
			\vspace{-8mm}
			\begin{IEEEbiography}[{\includegraphics[width=1in,height=1.25in,clip,keepaspectratio]{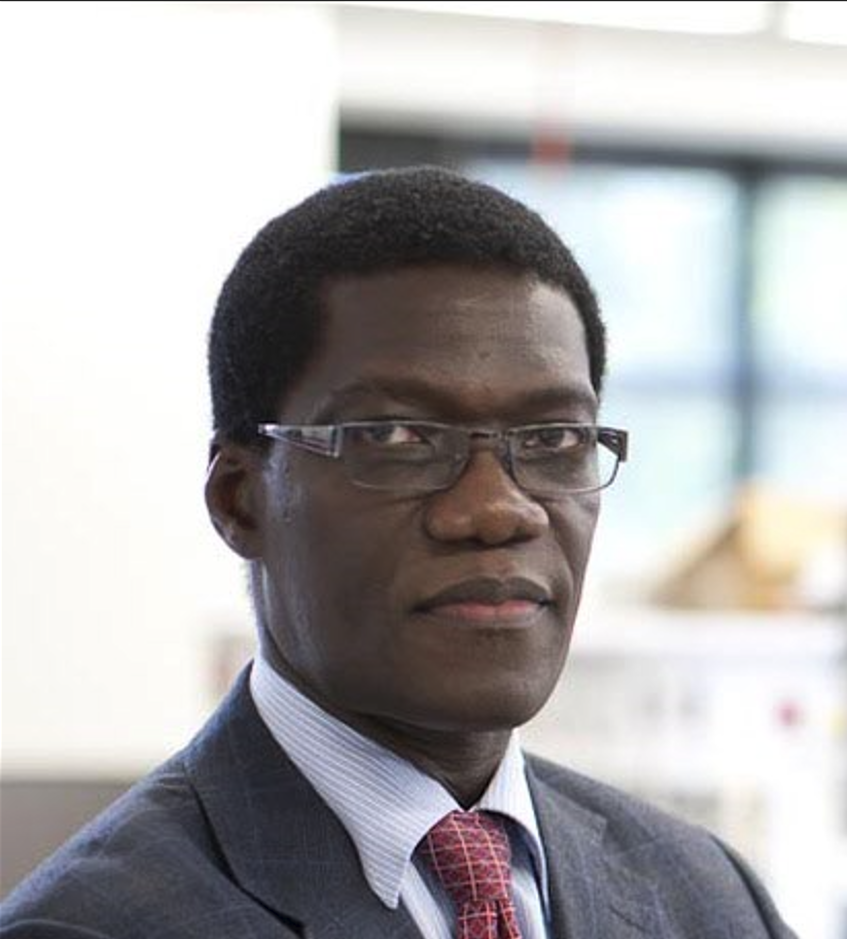}}]{Philip O. Ogunbona }  received the B.Sc. degree (with first class honours) in electronics and electrical engineering from the University of Ife, Nigeria, and the Ph.D. degree in electrical engineering from Imperial College London, U.K. He is a professor in computer science at the University of Wollongong, Australia. His research interests include signal and image processing, machine learning, computer vision and natural language processing. Professor Ogunbona is Fellow of the Australian Computer Society and a Life Senior Member of IEEE.\end{IEEEbiography}
			\vskip 0pt plus -1fil
			\vspace{-8mm}
			\begin{IEEEbiography}[{\includegraphics[width=1in,height=1.25in,clip,keepaspectratio]{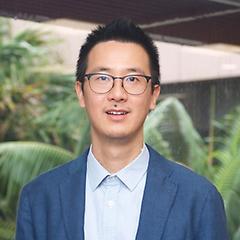}}]{Jie (Jack) Yang } is a lecturer of Big Data Analytics in the School of Computing and Information Technology at the University of Wollongong. His major research interests include Data Mining and Natural Language Processing. Dr. Yang has been the Chief Investigator for three research grants of Discovery/Linkage Project themes from the prestigious Australian Research Council (ARC) and has published over 70 articles.\end{IEEEbiography}

			\clearpage
			
			\twocolumn[
			\begin{center}
				{\LARGE\bfseries Supplementary Materials}
			\end{center}
			\vspace{1em}
			]
			
			\appendices


			%
			
			%
			%
			%
		%
		\maketitle              
		%

		%
		%
		%

		\section{Theoretical Analysis}
		
		We provide a theoretical analysis showing that diffusion enables noise-dependent learning of motion modes, allowing simultaneous modeling of subtle and strong motion.

		In masked autoencoders (MAE)~\cite{mae}, masked coordinates are directly regressed from visible ones by minimizing an MSE loss:
		\begin{equation}
			\mathcal{L}_{\text{MAE}}
			=
			\mathbb{E}\big[\|x_0^m - f_\theta(x_0^v)\|_2^2\big]
			=
			\sum_{i\in \text{masked}} (x_{0,i}-\hat{x}_{0,i})^2,
		\end{equation}
		where $x_0^m$ and $x_0^v$ denote the masked and visible coordinates, $f_\theta$ is the prediction model, and $\hat{x}_{0,i}$ is the reconstructed value of the $i$th coordinate.
		
		Even when $f_\theta$ is optimal, the gradient scale driving learning is proportional to the reconstruction residual:
		\begin{equation}
			\left\|
			\nabla_\theta (x_{0,i}-\hat{x}_{0,i})^2
			\right\|
			=
			2|x_{0,i}-\hat{x}_{0,i}|
			\left|
			\frac{\partial \hat{x}_{0,i}}{\partial \theta}
			\right|.
		\end{equation}
		Thus, coordinates associated with strong motions yield larger gradients and dominate optimization, while subtle motions contribute less without explicit reweighting.
		
		\medskip
		
		In contrast, the proposed framework performs diffusion in the latent embedding space while supervising denoising in coordinate-space motion. Let $S \in \mathbb{R}^{T_{in}\times V \times C_{in}}$ denote the input skeleton sequence. After temporal segmentation and embedding via $g_1(\cdot)$, the latent representation is obtained as
		\begin{equation}
			x = g_1(S), \qquad x \in \mathbb{R}^{\tau \times V \times C}.
		\end{equation}
		With spatial and temporal positional encodings, the clean latent sample is defined as
		\begin{equation}
			x_0 = x + pe^s + pe^t,
		\end{equation}
		where \(pe^s \in \mathbb{R}^{1 \times V \times C}\) and \(pe^t \in \mathbb{R}^{\tau \times 1 \times C}\).
		A masking operation partitions $x_0$ into visible and masked subsets, denoted by $x_0^v$ and $x_0^m$, respectively. 
		
		
		\medskip
		
		In the forward diffusion process, only the masked tokens, $x_0^m$, are corrupted by Gaussian noise. However, for notational clarity, we denote the clean latent sample as $x_0$
		\begin{equation}
			x_t = \sqrt{\bar{\alpha}_t}\,x_0 + \sqrt{1-\bar{\alpha}_t}\,\epsilon,
			\qquad
			\epsilon \sim \mathcal{N}(0,I),
		\end{equation}
		where $\bar{\alpha}_t$ controls the noise level at timestep $t$. The corresponding signal-to-noise ratio (SNR) is
		\begin{equation}
			\text{SNR}_t = \frac{\bar{\alpha}_t}{1-\bar{\alpha}_t}.
		\end{equation}
		Early timesteps correspond to high SNR (low noise), while later timesteps correspond to low SNR (high noise).
		
		\medskip
		
		The denoising decoder learns the conditional distribution $p(x_0^m \mid x_t^m, x_0^v)$ using $u$ as conditioning. The full denoising function is expressed as
		\begin{equation}
			\widehat{\dot{x}}_0 = D_{\text{Denoise}}(x_t^m, t, u),
		\end{equation}
		which predicts the motion of the clean sequence. Let the coordinate motion be defined by the temporal finite-difference operator
		\begin{equation}
			\dot{x}_0 = D x_0, \qquad Dx = [x^{\tau+1}-x^\tau].
		\end{equation}
		
		The training objective supervises motion reconstruction:
		\begin{equation}
			\mathcal{L}_{\text{denoise}}
			=
			\mathbb{E}\big[\|\dot{x}_0 - \widehat{\dot{x}}_0\|_2^2\big].
		\end{equation}
		
		\medskip
		
		To analyze the denoising behavior, assume the clean latent tokens follow a Gaussian distribution:
		\begin{equation}
			x_0 \sim \mathcal{N}(\mu_x,\Sigma_x).
		\end{equation}
		The posterior mean of the clean sample given $x_t$ is
		\begin{equation}
			\mathbb{E}[x_0 \mid x_t]
			=
			\mu_x + K_t (x_t - \sqrt{\bar{\alpha}_t}\mu_x),
		\end{equation}
		where
		\begin{equation}
			K_t =
			\sqrt{\bar{\alpha}_t}\,\Sigma_x
			\big(\bar{\alpha}_t\Sigma_x + (1-\bar{\alpha}_t)I\big)^{-1}.
		\end{equation}
		
		\medskip
		
		To align the analysis with the motion-based supervision, we define the latent motion variable
		\begin{equation}
			m_0 = D x_0.
		\end{equation}
		Since $D$ is linear, $m_0$ is Gaussian:
		\begin{equation}
			m_0 \sim \mathcal{N}(D\mu_x,\Sigma_m),
		\end{equation}
		with motion covariance
		\begin{equation}
			\Sigma_m = D\Sigma_x D^\top.
		\end{equation}
		The eigenvalues of $\Sigma_m$ quantify motion strength: large eigenvalues correspond to strong motion modes, while small eigenvalues correspond to subtle motion modes.
		
		Because $D$ is linear, the posterior mean of latent motion satisfies
		\begin{equation}
			\mathbb{E}[m_0 \mid x_t]
			=
			D\,\mathbb{E}[x_0 \mid x_t]
			=
			D\mu_x + D K_t (x_t - \sqrt{\bar{\alpha}_t}\mu_x).
		\end{equation}
		
		The diffusion gain along a motion mode with eigenvalue $\lambda$ is governed by
		\begin{equation}
			k_t(\lambda)
			=
			\frac{\sqrt{\bar{\alpha}_t}\lambda}
			{\bar{\alpha}_t\lambda + (1-\bar{\alpha}_t)}.
		\end{equation}
		
		For strong motion modes satisfying
		\begin{equation}
			\lambda \gg \frac{1-\bar{\alpha}_t}{\bar{\alpha}_t},
		\end{equation}
		the gain simplifies to
		\begin{equation}
			k_t(\lambda) \approx \frac{1}{\sqrt{\bar{\alpha}_t}},
		\end{equation}
		indicating that strong motion modes remain identifiable even at high noise levels.
		
		For subtle motion modes satisfying
		\begin{equation}
			\lambda \ll \frac{1-\bar{\alpha}_t}{\bar{\alpha}_t},
		\end{equation}
		the gain becomes
		\begin{equation}
			k_t(\lambda) \approx
			\frac{\sqrt{\bar{\alpha}_t}\lambda}{1-\bar{\alpha}_t},
			\qquad
			\text{as } \bar{\alpha}_t \to 1.
		\end{equation}
		Thus, subtle motion modes are most recoverable at early timesteps with high SNR.
		
		\medskip
		
		Finally, we relate latent motion denoising to coordinate-space supervision. Using a first-order approximation of the decoder around a reference point $\bar{x}$,
		\begin{equation}
			h_\psi(x) \approx h_\psi(\bar{x}) + J_h (x - \bar{x}),
		\end{equation}
		where $J_h$ is the Jacobian of the decoder. The predicted motion can be locally written as
		\begin{equation}
			\widehat{\dot{x}}_0 \approx c + J_h x_0,
		\end{equation}
		where $c$ is a constant. Substituting the posterior mean yields
		\begin{equation}
			\mathbb{E}[\dot{x}_0 \mid x_t]
			\approx
			c + J_h\,\mathbb{E}[m_0 \mid x_t].
		\end{equation}
		
		This shows that coordinate motion prediction inherits the same SNR-dependent structure as latent motion denoising. Therefore, provided that the decoder preserves motion-relevant directions, the model learns different motion scales depending on the noise level.
		
		\medskip
		
		Consequently, early diffusion steps with high SNR emphasize subtle motion modes, while later steps increasingly capture strong motion dynamics. Unlike direct masked regression, which is biased toward large residuals, latent diffusion with motion supervision distributes learning across the full spectrum of motion dynamics in a principled manner.

		\section{Additional Results}
		
		More experimental results are provided in this document.
		\subsection{Dataset}
		
		\subsubsection{NW-UCLA~\cite{NUCLA}} dataset consists of 1,494 action sequences performed by 10 subjects, recorded from three different viewpoints, and spanning 10 distinct action categories.

		\subsection{More Comparison Results}
		
		\subsubsection{Transfer Learning}
		
		The encoder and mapping layer are pre-trained on the NTU RGB+D~\cite{shahroudy01} 60 and NTU RGB+D 120~\cite{Liu1905} datasets separately using X-sub settings and are evaluated on the UCLA dataset using the supervised fine-tuning evaluation setting. The results are presented in Table~\ref{tab6_331}.   DiMoP outperforms TAHAR~\cite{TAHAR}  by 1.1 and 1.5 percentage points on the UCLA with pretraining on NTU RGB+D 60 and 120 datasets respectively.

		\begin{table}[thb]
			\caption{ Transfer learning evaluation of DiMoP on the UCLA dataset: Fine-tuning the pre-trained encoder from NTU RGB+D 60 and NTU RGB+D 120.}  
			\label{tab6_331}
			
			\centering
			\begin{tabular}{@{}l |cc}
				\cline{1-3} 
				Models    &   \multicolumn{2}{|c}{ To UCLA (\%)} \\
				& NTU 60 &  NTU 120 \\
				\cline{1-3} 
				TAHAR~\cite{TAHAR}       & 93.1  & 95.2 \\
				\cline{1-3} 
				\textbf{DiMoP (ours) }             &  \textbf{94.2}  &  \textbf{96.7} \\
				\cline{1-3} 
			\end{tabular}
			
		\end{table}

		\subsection{More Ablation Results}
		
		\subsubsection{Analysis of decoder configuration:}
		
		To further verify the source of DiMoP’s improvement, MacDiff’s~\cite{MacDiff} AdaLN-based conditional U-Net decoder was substituted into DiMoP while keeping the encoder, noise schedule, and training settings identical ($T=300$, linear schedule). The linear evaluation results are summarized in Table~\ref{tab_macdiff_decoder}. No performance gain was observed; accuracy slightly decreased under the same noise-prediction objective, and remained nearly unchanged when the motion-prediction target was applied. These outcomes indicate that DiMoP’s advantage arises primarily from its partial-state diffusion mechanism and motion-prediction objective, rather than from architectural or capacity differences in the decoder.
		
		\begin{table}[htbp]
			\centering
			\caption[Comparison with MacDiff decoder.]{Linear Evaluation Results on the NTU RGB+D 60 X-Sub Dataset After Replacing DiMoP’s Decoder with MacDiff’s AdaLN-Based Decoder.}
			\label{tab_macdiff_decoder}
			\begin{tabular}{l|c}
				\hline
				Model & Acc(\%) \\
				\hline
				MacDiff (reproduced) & 86.03 \\
				DiMoP (ours) & \textbf{86.76} \\
				DiMoP + MacDiff decoder (Noise) & 85.83 \\
				DiMoP + MacDiff decoder (Motion) & 86.61 \\
				\hline
			\end{tabular}
		\end{table}

		\subsubsection{Analysis of training stability:}
\begin{figure*}
			\centering
			\begin{subfigure}{0.45\textwidth}
				\centering
				\includegraphics[width=\linewidth]{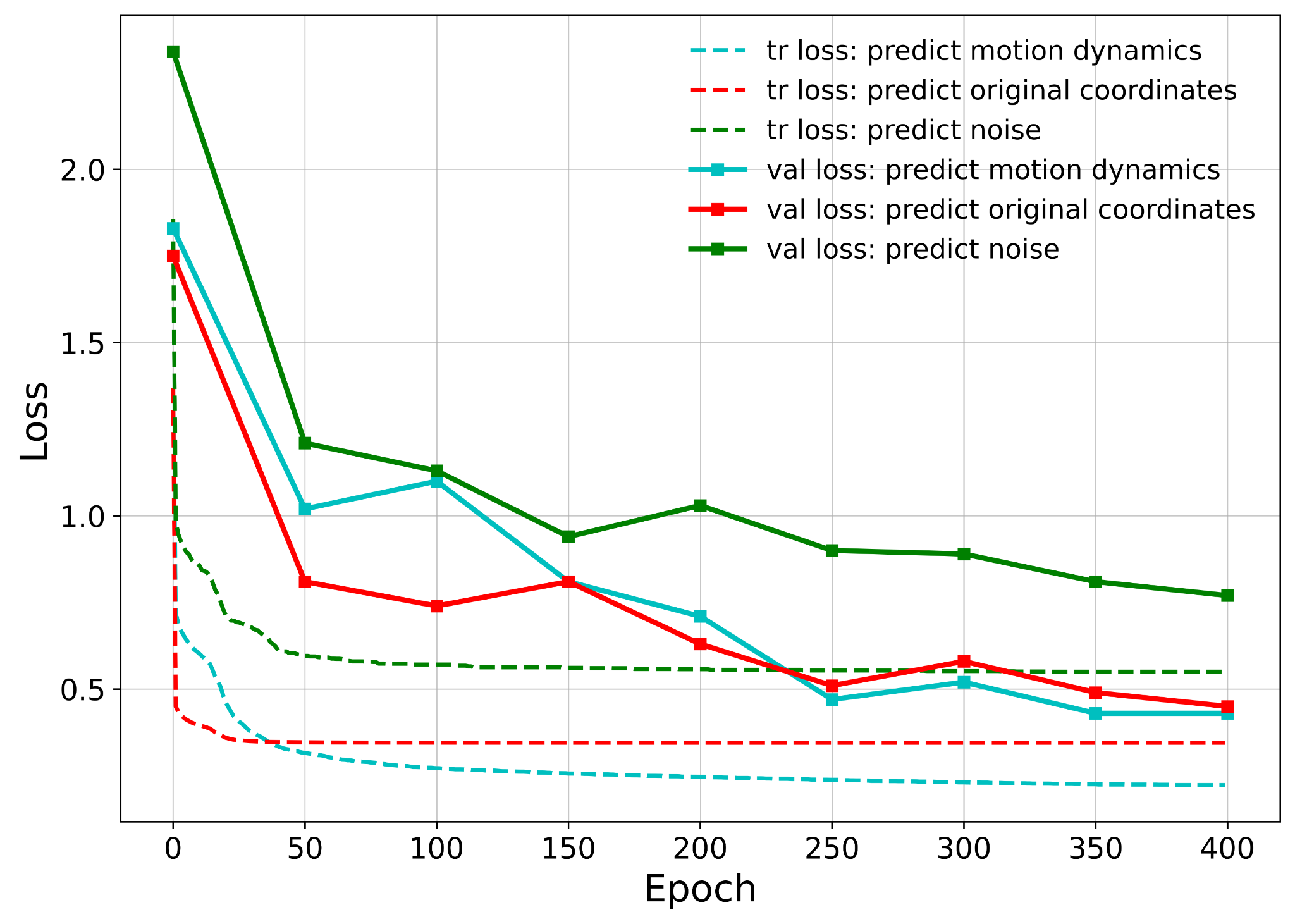}
				\caption{Varying the prediction target.}
				\label{fig:sub1}
			\end{subfigure}
			\hfill
			\begin{subfigure}{0.45\textwidth}
				\centering
				\includegraphics[width=\linewidth]{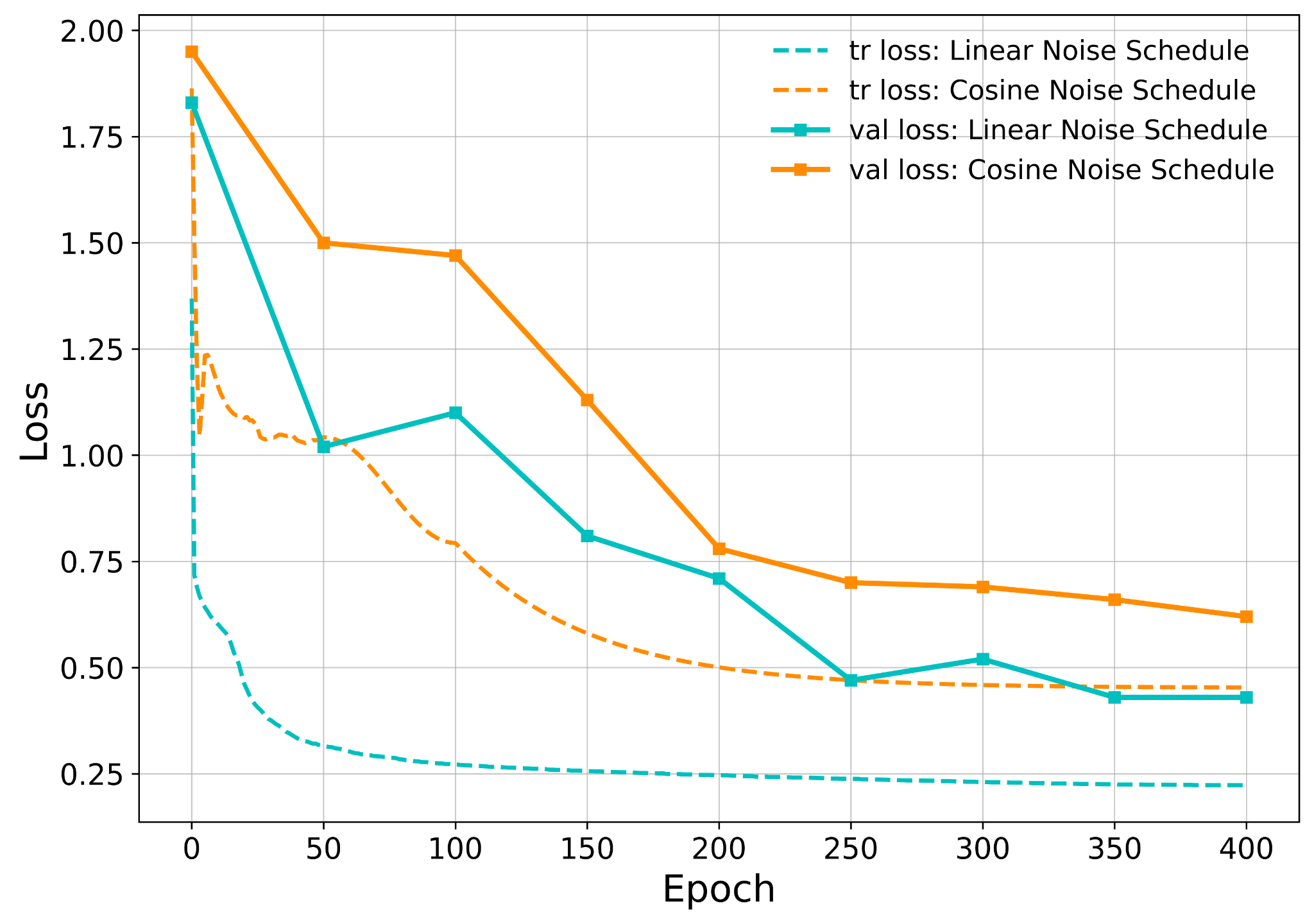}
				\caption{Varying the noise scheduling.}
				\label{fig:sub2}
			\end{subfigure}
			\hfill
			\begin{subfigure}{0.45\textwidth}
				\centering
				\includegraphics[width=\linewidth]{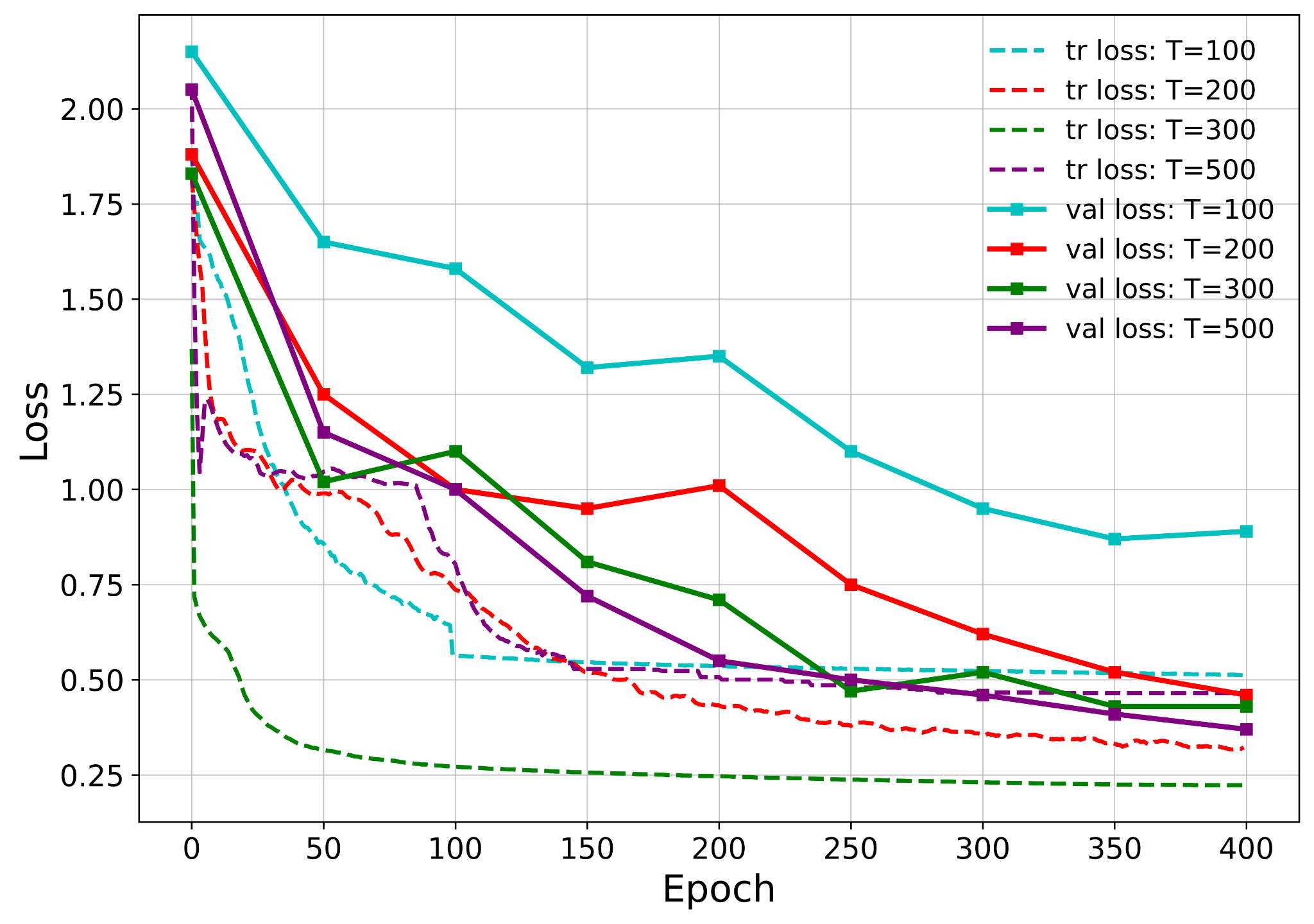}
				\caption{Varying the Diffusion steps $T$.}
				\label{fig:sub3}
			\end{subfigure}
			
			\caption{Training and validation loss curves obtained by varying the Diffusion hyperparameters.}
			\label{loss_curve_stability}
		\end{figure*}

		Training stability was evaluated under varying diffusion hyperparameter settings, including prediction targets, noise schedules, and diffusion step counts. During pre-training, validation was performed every 50 epochs on a held-out set, and both training and validation losses were recorded. The corresponding loss curves are shown in Figure~\ref{loss_curve_stability}. In all cases, convergence of both training and validation losses was observed, demonstrating that stable optimization was achieved and robustness to different diffusion configurations was maintained.
		
		\subsubsection{Analysis of computational complexity:}
		We further analyze the computational overhead of DiMoP compared to the baseline MAMP~\cite{mao01}. During pre-training on NTU RGB+D 60 (X-View) with four NVIDIA P100 GPUs, DiMoP incurs slightly higher complexity due to the additional mapping MLP, pseudo-classifier, and denoising decoder. As shown in Table~\ref{tab:cost_comparison}, the parameter count increases from 8.79M to 9.69M and FLOPs from 5.45G to 6.78G. Importantly, these components are only used in pre-training and are not involved during downstream inference.
		
		For downstream tasks, only the encoder and mapping layer are required. Table~\ref{tab:cost_comparison} reports the parameter count and inference time per iteration for linear evaluation on two P100 GPUs. DiMoP requires 0.38M parameters and 0.413s per iteration, compared to 0.27M and 0.342s for MAMP. The overhead is marginal and remains practically manageable, while offering consistent accuracy improvements.
		
		\begin{table}
			\centering
			\caption{Computational cost comparison of MAMP and DiMoP during pre-training and linear evaluation.}
			\label{tab:cost_comparison}
			\begin{tabular}{lcc|cc}
				\hline
				& \multicolumn{2}{c|}{Pre-training} & \multicolumn{2}{c}{Linear Evaluation} \\
				\hline
				Model & Params & FLOPs (G) & Params & Time (s/iter) \\
				\hline
				MAMP~\cite{mao01} (baseline) & 8.79M & 5.45 & 0.27M & 0.342 \\
				DiMoP & 9.69M & 6.78 & 0.38M & 0.413 \\
				\hline
			\end{tabular}
		\end{table}

		\subsubsection{Statistical Significance Analysis:}
		
		To further assess statistical significance, the linear evaluation was repeated five times for both MAMP and DiMoP on NTU RGB+D 60 X-sub and X-view. In each trial, a linear classifier was trained on top of the frozen encoder with random initialization. A paired two-tailed $t$-test was then applied at $\alpha=0.05$. The null hypothesis was defined as no significant difference between the two methods, i.e., $H_0: \mu_{\text{DiMoP}}=\mu_{\text{MAMP}}$, while the alternative hypothesis was defined as a significant difference, i.e., $H_1: \mu_{\text{DiMoP}}\neq\mu_{\text{MAMP}}$.
		
		\begin{table}[h]
			\centering
			\caption{Paired two-tailed $t$-test results between MAMP and DiMoP on NTU RGB+D 60.}
			\label{tab:supp_ttest}
			\small
			\begin{tabular}{lccccc}
				\hline
				Protocol & $\mu_{\text{MAMP}}$ & $\mu_{\text{DiMoP}}$ & $t$-value & $t_{\text{crit}}$ & Decision \\
				\hline
				X-sub  & 84.95 & 86.79 & 5.636 & 2.776 & Reject $H_0$ \\
				X-view & 89.14 & 91.91 & 8.154 & 2.776 & Reject $H_0$ \\
				\hline
			\end{tabular}
		\end{table}
		
		With four degrees of freedom, the computed $t$-values exceed the critical value for both protocols. Therefore, $H_0$ is rejected, indicating that the performance difference between DiMoP and MAMP is statistically significant. Since DiMoP obtains higher mean accuracy in both protocols, the significant difference supports the superiority of DiMoP over MAMP under this evaluation.
		
		\subsubsection{Reproducibility of Baseline Results:}
		
		The two most closely related publicly available baselines, MAMP~\cite{mao01} and MacDiff~\cite{MacDiff}, were reproduced using their released implementations. As shown in Table~\ref{tab:supp_reproduced_baselines}, the reproduced linear evaluation results are highly consistent with the originally reported values, with only minor deviations caused by training stochasticity.
		
		\begin{table}[h]
			\centering
			\caption{Linear evaluation performance of reproduced baselines. $^{*}$ denotes reproduced results.}
			\label{tab:supp_reproduced_baselines}
			\small
			\begin{tabular}{lccccc}
				\hline
				\multirow{2}{*}{Model} & \multicolumn{2}{c}{NTU 60} & \multicolumn{2}{c}{NTU 120} & PKU \\
				\cline{2-6}
				& X-sub & X-view & X-sub & X-set & Part I \\
				\hline
				MAMP & 84.90 & 89.10 & 78.60 & 79.10 & 92.20 \\
				MAMP$^{*}$ & 84.84 & 89.13 & 78.49 & 78.98 & 92.24 \\
				MacDiff & 86.40 & 91.00 & 79.40 & 80.20 & 92.80 \\
				MacDiff$^{*}$ & 86.43 & 90.95 & 79.34 & 80.13 & 92.74 \\
				\hline
			\end{tabular}
		\end{table}
		
		For semi-supervised evaluation, the exact MacDiff configuration was not provided with the released code. Therefore, MacDiff was re-evaluated under the same data split, label ratio, training setting, and evaluation protocol used for DiMoP to ensure a controlled comparison.
		
		\subsubsection{Performance on actions with subtle vs strong Motion:}
		
		
		
		Table~\ref{tab:subtle_strong} provides a group-level interpretation of the class-wise results. The larger gain on subtle-motion actions suggests that DiMoP better preserves localized, low-amplitude cues that are easily dominated by stronger body movements in direct masked reconstruction. This indicates a key limitation of MAE-style objectives: reconstruction can be biased toward high-energy motion components, whereas progressive denoising exposes the model to fine deviations around the clean motion manifold at high-SNR stages. The gain on strong-motion actions further shows that DiMoP does not improve subtle actions at the expense of large-scale dynamics, but also retains global displacement, posture transitions, and interaction patterns through later denoising stages. Therefore, the observed improvements are not merely class-wise fluctuations; they indicate that DiMoP learns motion representations across multiple spatial and temporal scales.
		
		\begin{table}
			\centering
			\caption{Group-wise accuracy improvement of DiMoP over MAMP on NTU RGB+D 60 X-sub.}
			\label{tab:subtle_strong}
			\small
			\begin{tabular}{l | p{0.42\columnwidth} | c}
				\hline
				Group & Dominant motion cue & Average \% Gain \\
				\hline
				Subtle motion & Localized low-amplitude hand, arm, head, and object-related motion & +5.77 \\
				\hline
				Strong motion & Global displacement, posture transition, and interaction dynamics & +4.31 \\
				\hline
			\end{tabular}
			
			\vspace{1mm}
			\footnotesize
			Subtle: A2, A3, A11, A12, A14, A16, A17, A29, A30, A31, A33, A34, A36, A41, A44, A45, A46.
			Strong: A6, A7, A8, A9, A15, A20, A22, A24, A26, A27, A40, A50, A51, A52, A55, A57, A58.
		\end{table}
		
\begin{figure}
			\centering
			\includegraphics[width=\columnwidth]{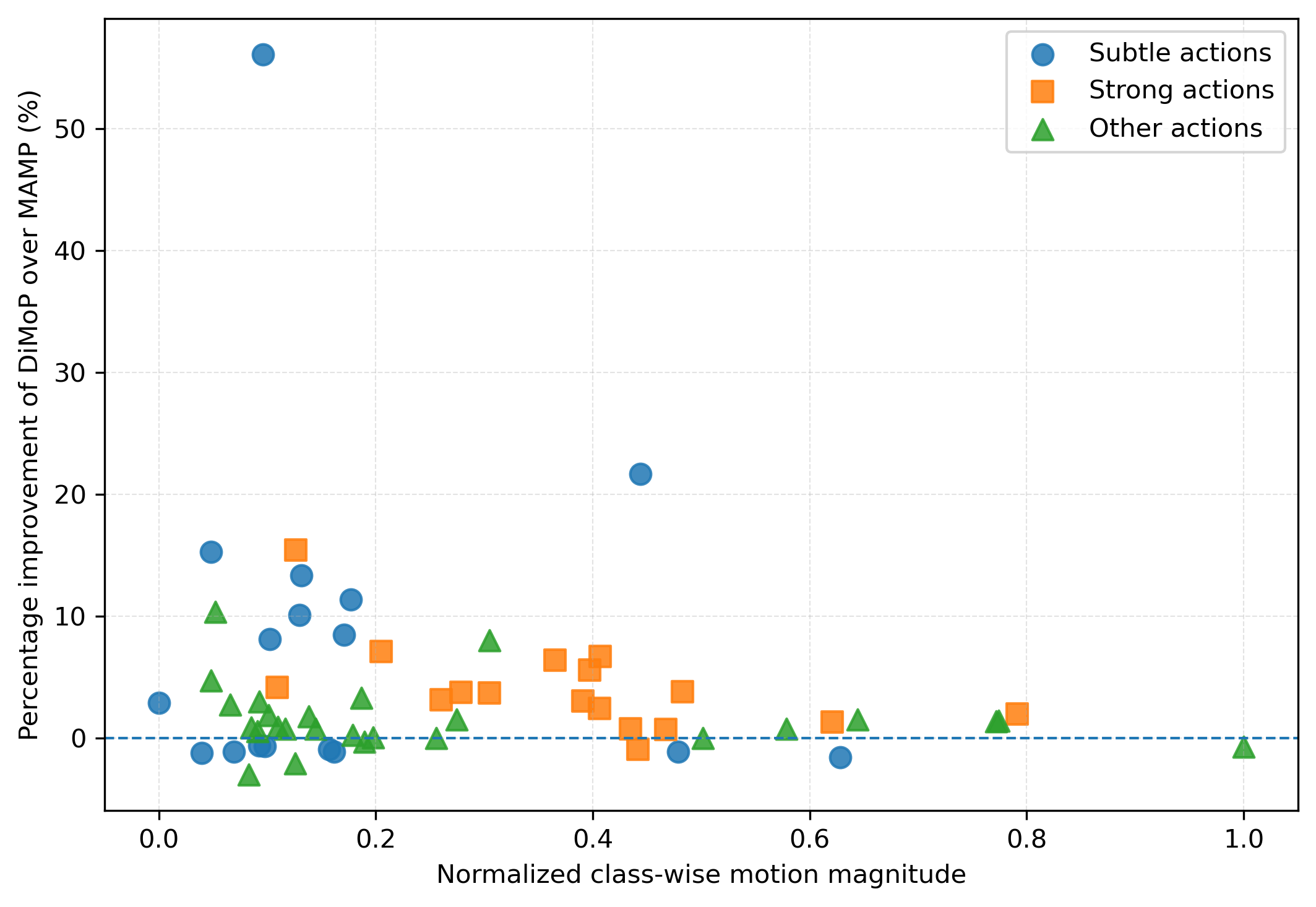}
			\caption{Class-wise normalized motion magnitude versus relative percentage improvement of the actions Subtle and strong motions with DiMoP over MAMP on NTU RGB+D 60 X-sub.}
			\label{fig:motion_gain_scatter}
		\end{figure}

		To further examine whether the improvements are biased toward high-motion classes, class-wise motion magnitude was computed as the average frame-wise joint displacement normalized by body size. The body size was estimated using the spine-to-neck bone length, which provides a stable torso-based scale factor and reduces the influence of inter-subject size variation. Fig.~\ref{fig:motion_gain_scatter} plots the normalized motion magnitude against the relative percentage improvement of DiMoP over MAMP.
		
		The scatter distribution shows that DiMoP improves classes across a broad range of normalized motion magnitudes, including several low-motion subtle actions. This indicates that the proposed diffusion-driven learning does not merely benefit high-amplitude actions, but also enhances weak and fine-grained motion representations. Together with the group-wise analysis, these results suggest that DiMoP functions as a multi-scale motion learner: high-SNR denoising stages preserve subtle local deviations, while later stages maintain robustness to large-scale temporal and spatial dynamics.
		
		\begin{figure}
			\centering
			\includegraphics[width=\columnwidth]{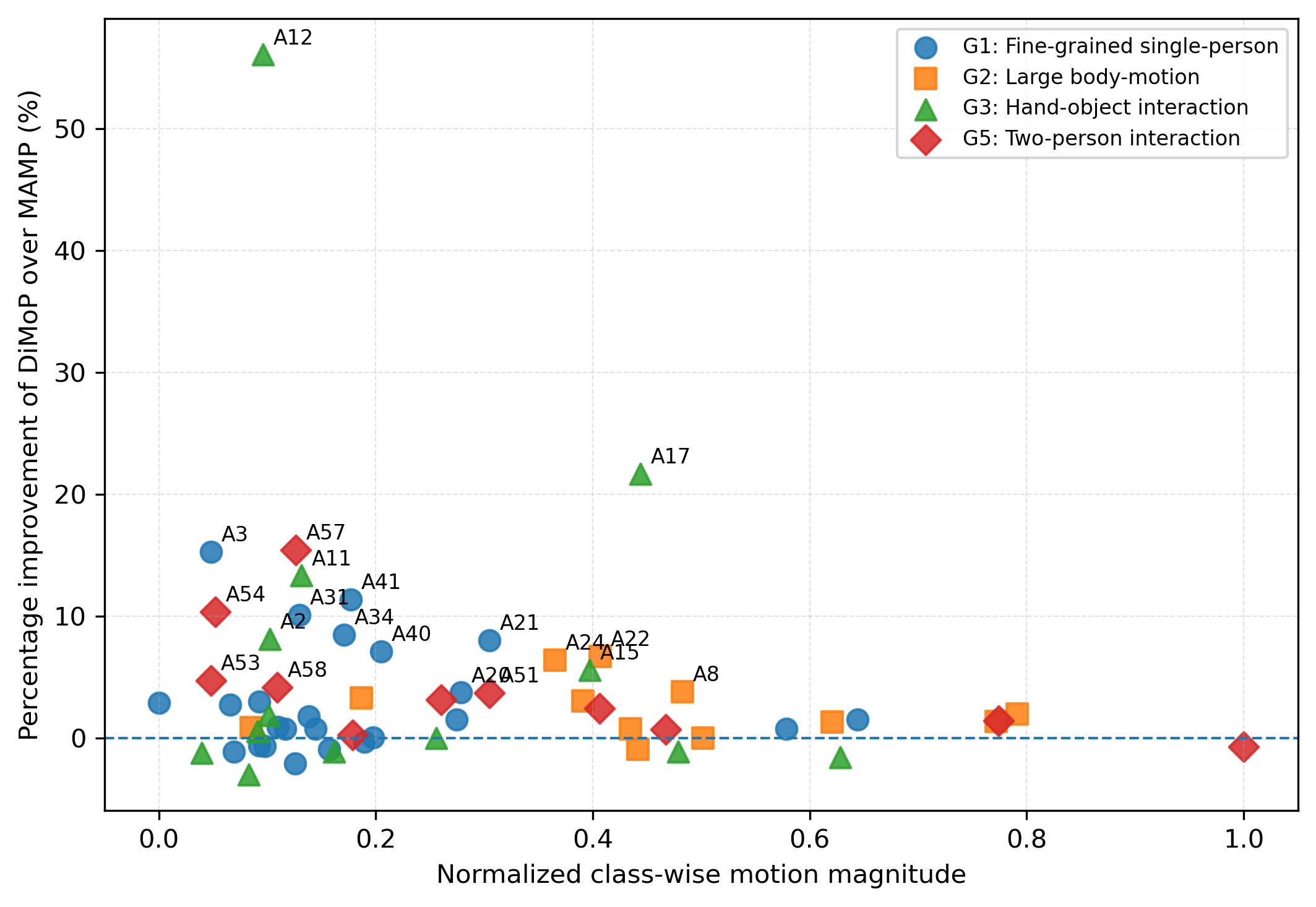}
			\caption {Group level analysis of normalized class-wise motion magnitude against the relative percentage improvement of DiMoP over MAMP on NTU60 X-sub dataset under linear evaluation. }
			\label{fig3}
		\end{figure}

		\subsubsection{Group-wise and Motion-magnitude Analysis:}
		
		We analyze how DiMoP improves performance on actions with different characteristics in comparison with the baseline MAMP. In particular, the 60 actions are divided into 5 groups (non-exclusively): G1 - Fine-grained single-person actions; G2 - Large body-motion actions; G3 - Hand-object interaction actions; G4 - Two-person interaction actions; G5 - Weak-discriminative-frame actions. Please note that the G5 overlaps with other groups, as some actions share multiple characteristics. 
		
		\begin{table*}
			\centering
			\caption{Performance improvement on different groups of actions under linear evaluation between DiMoP and MAMP on the NTU RGB+D 60 X-sub dataset and the analysis of probable contributing factors provided by the proposed method.}
			\label{tab:supp_groupwise}
			\small
			\begin{tabular}{p{0.19\textwidth} p{0.27\textwidth} c c p{0.29\textwidth}}
				\hline
				Group & Action IDs & 
				\makecell{Improved/Dropped\\actions (\% Improved)} & 
				\makecell{Mean\\$\Delta$Acc.} & 
				Main Insight \\
				\hline
				G1: Fine-grained single-person actions &
				A3, A4, A18--A21, A28, A31, A33--A41, A44--A49 &
				17/6(74\%) & +1.88  &
				Localized hand, arm, head, and object-related motions were better captured, while drops mainly occurred when cues were confined to similar body regions. \\
				
				G2: Large body-motion actions &
				A5--A9, A22--A24, A26, A27, A42, A43 &
				11/1 (92\%) & +2.19 &
				Global displacement, posture transitions, and strong temporal dynamics were consistently preserved. \\
				
				G3: Hand-object interaction actions &
				A1, A2, A10--A17, A25, A29, A30, A32 &
				9/5(64\%) & +2.59 &
				Local-to-mid-level hand motion was improved, although skeleton-only input remained limited when object/contact cues were essential. \\

				G4: Two-person interaction actions &
				A50--A60 &
				10/1 (91\%)& +3.57 &
				Interaction-related motion cues were effectively captured, although explicit inter-subject relation modeling may further improve performance. \\

				G5: Brief-discriminative-frame actions$^{\dagger}$ &
				A3, A18--A21, A25, A28, A33, A37, A41, A44--A48, A57 &
				10/6(63\%) & +2.16 &
				Actions with short-lived discriminative cues remained challenging when most frames contained neutral or visually similar poses. \\

				Overall &
				A1--A60 &
				47/13 (78\%)& +1.90 &
				Most classes were improved, supporting the effectiveness of diffusion-based motion distribution learning. \\
				\hline
			\end{tabular}
			
			\vspace{1mm}
			\footnotesize
			$^{\dagger}$The brief-discriminative-frame group overlaps with other semantic groups and is used only to analyze temporal failure cases.
		\end{table*}
		
		As shown in Table~\ref{tab:supp_groupwise}, positive average gains were obtained across all motion groups, indicating that the improvements were not restricted to isolated classes. Overall, DiMoP improves 47/60 (78\%) classes and achieves positive average gains (1.9 percentage points) across all groups of actions. The gains are more evident in fine-grained actions, hand-object interaction actions, and two-person interaction actions, with 74\%, 92\%, and 91\% of actions improved, respectively. This supports the claim that progressive denoising captures both localized and interaction-related motion dynamics. As expected, the two types of actions, hand-object interaction and weak discriminative frame, have the lowest percentage of the actions being improved.  Furthermore, the  Figure~\ref{fig3} presents the normalized class-wise motion magnitude against the relative percentage improvement of DiMoP over MAMP.

		\subsubsection{Effect of Increasing the Temporal Smoothness Weight:}
		
		To examine whether the value consistency loss can be replaced by a larger temporal smoothness weight, we conducted an additional ablation on NTU RGB+D 60 X-sub under linear evaluation. The denoising loss was kept unchanged, while $\lambda$ was increased without using $\mathcal{L}_{consistency}$.
		
		\begin{table}[h]
			\centering
			\caption{Effect of increasing $\lambda$ without $\mathcal{L}_{consistency}$ on NTU RGB+D 60 X-sub under linear evaluation.}
			\label{tab:supp_smooth_consistency}
			\small
			\begin{tabular}{c c c c}
				\hline
				$\alpha$ & $\lambda$ & $\gamma$ & Acc. (\%) \\
				\hline
				1 & 0.3 & 0   & 85.33 \\
				1 & 1.0 & 0   & 84.92 \\
				1 & 3.0 & 0   & 84.59 \\
				1 & 5.0 & 0   & 83.46 \\
				1 & 0.3 & 1.0 & 86.48 \\
				\hline
			\end{tabular}
		\end{table}
		
		As shown in Table~\ref{tab:supp_smooth_consistency}, increasing $\lambda$ without $\mathcal{L}_{consistency}$ does not recover the performance of the full objective. Instead, large $\lambda$ values progressively degrade accuracy, indicating that excessive temporal smoothing over-regularizes the pseudo-classifier and weakens representation learning. This confirms that $\mathcal{L}_{smooth}$ and $\mathcal{L}_{consistency}$ are complementary: $\mathcal{L}_{smooth}$ regularizes local frame-to-frame transitions, whereas $\mathcal{L}_{consistency}$ imposes global sequence-level agreement.

		%
		%
		%

\end{document}